\documentclass{article}
\usepackage[preprint]{neurips_2026}
\usepackage[T1]{fontenc}
\usepackage[utf8]{inputenc}
\usepackage{lmodern,microtype}
\usepackage{amsmath,amssymb,graphicx,booktabs,tabularx,array,float,xcolor}
\usepackage{amsthm}
\usepackage{algorithm}
\usepackage{algpseudocode}

\usepackage{tikz}
\usetikzlibrary{arrows.meta,decorations.pathreplacing,patterns}

\let\RtwoOriginalIncludeGraphics\includegraphics
\renewcommand{\includegraphics}[2][]{%
  \IfFileExists{#2}{%
    \RtwoOriginalIncludeGraphics[#1]{#2}%
  }{%
    \fbox{\parbox[c][28mm][c]{0.88\linewidth}{%
      \centering
      \textcolor{red}{\textbf{Figure unavailable in this preview}}\\[3pt]
      {\small\texttt{\detokenize{#2}}}%
    }}%
  }%
}

\newcommand{\RtwoInputFigure}[1]{%
  \IfFileExists{#1}{%
    \input{#1}%
  }{%
    \PackageWarning{R2T2}{External figure omitted from preview: #1}%
    \fbox{\parbox[c][18mm][c]{0.88\linewidth}{%
      \centering\small
      External figure omitted from this preview.\\[2pt]
      {\scriptsize\texttt{\detokenize{#1}}}%
    }}%
  }%
}

\usepackage{xcolor}
\usepackage{float}
\usepackage{placeins}

\usepackage{soul}
\sethlcolor{yellow}

\newif\ifshowedits
\showeditsfalse

\ifshowedits

\else

\fi

\usepackage{hyperref}
\hypersetup{
    colorlinks=true,
    linkcolor=blue!55!black,
    citecolor=blue!55!black,
    urlcolor=blue!55!black
}

\newcommand{\qwen}{\textsc{Qwen3-ASR-1.7B}}
\newcommand{\method}{\textsc{R2T2}}
\newcommand{\waittoken}{\textnormal{\textsc{Wait}}}
\newcolumntype{L}[1]{>{\raggedright\arraybackslash}p{#1}}

\newtheorem{theorem}{Theorem}

\newtheorem{example}{Example}
\newtheorem{definition}{Definition}
\newtheorem{proposition}{Proposition}
\newcommand{\sysname}{Confucius4-R2T2}

\newcommand{\prefeq}{\preceq}
\newcommand{\lcp}{\mathrm{lcp}}
\newcommand{\LSP}{\mathrm{LSP}}

\title{Confucius4-R2T2: A Stable-Prefix Streaming ASR Model for
Real Real-Time Transcription}

\author{%
  Daniel Huang\setcounter{footnote}{1}\thanks{NetEase Youdao, Beijing, China. Corresponding author: \texttt{huangd@rd.netease.com}.},  Qi He\footnotemark[2], Shaoxiong Duan\setcounter{footnote}{2}\thanks{UC San Diego. \texttt{s4duan@ucsd.edu}.},  Yingjie Li\footnotemark[2], Qiao Cheng\footnotemark[2], Yitao Duan\footnotemark[2] \\%\thanks{Use footnote for providing further information
}

\date{}

\begin{document}
\maketitle

\begin{abstract}
Streaming speech recognition is increasingly consumed by machines, such as
translators and voice agents, that act on each token as it appears. A token
that is later revised costs wasted work or a wrong action. In this work, we present 
\sysname{} (short for Real Real-Time Transcription), a stable-prefix streaming ASR model 
that never modifies its earlier output while maintaining high accuracy and low latency. 
\method{} is obtained by post-training Qwen3-ASR-1.7B on 1,500 hours of Chinese 
and English audio while freezing its pre-trained audio encoder and projector.  The adaptation is 
based on the concept of \emph{Longest Stable Prefix} (LSP), a new notion that we 
introduce in this work. We prove that LSP is monotone under the prefix order, consistent with the full-audio 
output, and maximal among labels that equal a whole earlier output. 
Learning from LSP labels converts a full-context model into a stable-prefix
streaming one \emph{without} any architectural change. Extensive evaluations show 
that R2T2 brings streaming recognition close to the base model's
full-context quality while delivering stable text with low emission lag. Its
gains extend beyond the post-training languages, demonstrating a practical
path from full-context ASR to stable streaming. We open-source the model weights and code
at \url{https://huggingface.co/netease-youdao/Confucius4-R2T2} and 
\url{https://github.com/netease-youdao/Confucius4-R2T2}.

\end{abstract}

\section{Introduction}
\label{sec:introduction}

Streaming automatic speech recognition (ASR) must produce useful
transcriptions while audio is still arriving. The rapid advance of large
language models (LLMs) with strong reasoning and tool-use capabilities
increases the demand for partial transcripts to be consumed immediately by
something else: a live caption on a screen, a downstream machine translation
system, or a voice AI agent that must decide when to respond and which tool
to call. In all of these settings, an emitted token is an \emph{action}, in
the sense that downstream components start working on it. A well-documented
failure mode is that tokens emitted early are later revised as more audio
arrives, a phenomenon variously called \emph{flicker}
\citep{shangguan2020stability}, \emph{revision} \citep{niehues2018low}, or
\emph{hypothesis instability} \citep{bruguier2023flickering}. When the
recognizer revises a token, the consumer must discard or redo the work built
on it, and a human reader sees the text change. A revision is therefore not
a cosmetic defect. It is wasted computation and, in an agent, a possibly
wrong side effect. This irreversibility is why \emph{output stability},
meaning that earlier output is never modified by later output, matters as
much as accuracy and latency.

Stability has been recognized as a practical concern for some time, and prior
work addresses it from different perspectives and under different terms.
Some methods rely on inference-time commitment rules.
\citet{liu2020partialhypothesis} introduce \emph{local agreement}, which commits only the
longest common prefix of two consecutive partial hypotheses.
\citet{polak2022cuni} extend this to the beam level, call the committed
output a \emph{stable prefix}, and compare several commitment strategies
(Hold-$n$, LA-$n$, SP-$n$). \citet{arivazhagan2020retranslation} propose
\emph{target-suffix deletion} (mask-$k$), which discards the last $k$ tokens
of each partial hypothesis. A complementary line trains for stability
indirectly. \emph{Prefix training} \citep{niehues2016pre,niehues2018low}
exposes the model to randomly truncated training pairs, and self-training
\citep{sen2023selftraining} further improves the flicker--latency trade-off.
Both lower the revision rate, but neither guarantees that earlier tokens are
never revised. In simultaneous machine translation, \citet{hoang2026dynamic}
train on stable-prefix targets and obtain stability by construction through
prefix decoding.

In this work, we present \sysname{}, a streaming ASR model that treats output
stability as a first-class requirement. Before proceeding, we state
precisely what we require of such a model, because the word ``stable'' has
been used for different objects such as a statistic of a system's output over time
\citep{mcgraw2012wordstability}, an inference-time commit rule applied to an
unchanged model \citep{liu2020partialhypothesis}, or a training target
\citep{hoang2026dynamic}.

\begin{definition}[Stable-prefix streaming ASR model]
\label{def:spmodel}
Let $x_{\le t}$ be the audio received up to time $t$. A streaming ASR model
$\mathcal{M}$ is a recognizer together with its decoding strategy. 
Let $y_t = \mathcal{M}(x_{\le t})$ be its released text at time
$t$. $\mathcal{M}$ is a \emph{stable-prefix} streaming ASR model if, for every
audio signal $x$ and all $t \le t'$, $y_t \prefeq y_{t'}$, where $\prefeq$
denotes the prefix order.
\end{definition}

We highlight two key properties. First, it is purely \emph{behavioral}. It constrains only the ordering of
outputs, not mechanisms such as causal encoders or chunked attention, nor
training recipes. Such mechanisms are neither sufficient (a causal encoder
can feed a decoder that still revises) nor necessary (a bidirectional encoder
behind a non-retracting commit rule is stable). Second, stability is
\emph{orthogonal} to accuracy and latency. A system that emits nothing until
the end of the utterance is stable but slow, and one that emits a fixed token
sequence is stable but inaccurate. Accuracy and latency already have
well-calibrated metrics, so we treat stability as a \emph{constraint} to be
satisfied while optimizing them. These considerations motivate our definition of the longest stable prefix in Section \ref{sec:lsp_learning_main}, 
leading to some subtle but important differences from related concepts, and enabling an 
effective non-streaming-to-streaming adaptation method. 

\sysname{}, short for Real Real-Time Transcription, is a stable-prefix
streaming ASR model built by post-training Qwen3-ASR-1.7B on $1{,}500$ hours of
additional Chinese and English audio. It retains the pre-trained audio encoder
and its bidirectional processing within the available audio window, so the
encoder is neither causalized nor masked. \sysname{} supports configurable
chunk sizes from $80$\,ms to $2$\,s. Our main contributions are as follows:

\begin{enumerate}

  \item We introduce a new concept of Longest Stable Prefix (LSP) and prove 
  a number of its properties. We show that our definition provides the label that can 
  be extracted from a non-stable model to train a stable one. We
  prove that our label is monotone under the prefix order, consistent with the full-audio output, and
  maximal among labels that equal a whole earlier output.
  
  \item We present LSP learning, a two-stage training recipe that converts a full-context 
  model into a stable-prefix streaming model with LSP labels, without causalizing the 
  encoder or using attention masks.
  
  \item We open-source \sysname{} and conduct extensive evaluations on a range of benchmarks. 
  Our experiments show that \sysname{} achieves state-of-the-art performance in terms of recognition 
  accuracy (approaching full-context recognizers) and latency, even compared against proprietary models, 
  while maintaining a stable-prefix guarantee.
\end{enumerate}

\section{Related Work}
\label{sec:relate_work}

Streaming ASR trades off accuracy, emission latency, and output
stability. We organize prior work by which of these it controls.
Stability, meaning that emitted text is never revised, is a property of
the whole model, including its decoding and commit rule, not of the
encoder alone.

\paragraph{Architectures.}
Streaming architectures bound the acoustic context seen by the encoder,
through monotonic attention \citep{chiu2018mocha}, or through chunked,
cascaded, or dual-mode models that serve both streaming and
full-context operation
\citep{yu2021dualmode,zhang2021u2,wu2021u2plusplus,narayanan2021cascaded}.
Released systems include the streaming Zipformer transducer
X-ASR \citep{xasr_modelcard} and the cache-aware FastConformer--RNNT model Nemotron 3.5 \citep{nvidia2026nemotron35asr}.
Pre-trained full-context models have also been converted for streaming,
by adding a CTC branch with streaming attention masks
\citep{zhou2025twopasswhisper} or by causalizing the encoder
\citep{krichli2026whisperrt}.
These designs decide how context is processed. They do not by
themselves guarantee that emitted text is final.

\paragraph{Output stability.}
Work on stability falls into three groups.
\emph{Measurement:}  \citet{mcgraw2012wordstability} call a prefix stable if all
later incremental results share it, and estimate, per word, the
probability of this event so that a client can display the longest
prefix above a threshold;
\citet{shangguan2020stability} analyze instability of partial hypotheses.
\emph{Inference-time repair:} reranking favors consistency with
earlier output \citep{bruguier2023flickering}, while Local Agreement
and hold-$n$ commit a common prefix of successive hypotheses or
withhold their last $n$ tokens \citep{liu2020partialhypothesis}. The
same idea is used with frozen Whisper models
\citep{machacek2023whisperstreaming,wang2024simulwhisper}, and the
Qwen3-ASR-1.7B configuration in our comparisons re-decodes accumulated
audio and revises an unconfirmed suffix \citep{shi2026qwen3asr}.
These are per-step rules. Paired with a buffer that never retracts, they
yield non-retracting output, but the underlying recognizer keeps
revising and agreement is not learned.
\emph{Training:} in simultaneous translation, \citet{hoang2026dynamic}
train on targets anchored to the translation of the complete source,
keeping the committed length nondecreasing, and obtain flicker-free
output by decoding with the committed prefix forced.
Our target follows the same final-anchored principle.\footnote{We became
aware of \citet{hoang2026dynamic}, which was released very recently, after our method was developed; the
two lines of work were pursued independently.}
It differs in crediting a raw output only when the whole output is a
prefix of the full-audio output, and it is used in speech recognition.

\paragraph{Emission timing in speech language models.}
Speech ReaLLM defers generation with a BLANK token and
alignment-based training \citep{seide2024speechreallm}, SpeechLLM-XL
uses chunk-aligned targets from CTC alignment
\citep{jia2025efficientstreamingllm}, and Voxtral Realtime learns
delayed emission from timestamp-based targets with non-emitting padding
\citep{mistral2026voxtralrealtime}. Other work learns segmentation
policies or derives read/write decisions at inference from output
confidence \citep{wan2026decoderonlystreaming,guo2025readwritepolicy}.
These methods time emission against the acoustics. We instead time it
against what the full-context model will output.

\section{Preliminaries}
\label{sec:preliminaries}

\subsection{Notation}\label{sec:lsp-notation}
Let $x=(x_1,\dots,x_T)$ be an utterance of $T$ audio frames of $\delta=80$\,ms,
and $x_{\le t}$ its first $t$ frames. The recognizer $\mathcal{M}$ maps any
audio prefix to a token sequence. Write
\begin{align}
  \bar y &= \mathcal{M}(x), \label{eq:bar}\\
  y_t &= \mathcal{M}(x_{\le t}), \qquad n_t = |y_t|, \label{eq:yt}\\
  \ell_t &= \bigl|\lcp(y_t,\bar y)\bigr|, \label{eq:ell}
\end{align}
where $\operatorname{lcp}(\cdot)$ denotes the longest common prefix shared by all input sequences and
$\bar y$ is the \emph{full-context output}, not a human reference: it is
what the same model produces when it sees the whole utterance (so $y_T=\bar y$). We write $u\prefeq v$ if $u$ is a
prefix of $v$. In streaming operation the model is queried every $\Delta$, a
chunk length that is a multiple of $\delta$. At each query the encoder is
applied to the audio received so far, so the representations of $x_{\le t}$ may
differ from those of the first $t$ frames of the full utterance.

\subsection{Goals}
Besides the stable-prefix property (Definition~\ref{def:spmodel}), we aim for:
\begin{itemize}
  \item \textbf{High accuracy:} WER comparable to full-context ASR models of
  similar size.
  \item \textbf{Low latency:} below 500\,ms at the
  smallest chunk sizes.
  \item \textbf{Configurability:} the chunk length $\Delta$ is configurable to prioritize 
  either quality or speed, without retraining.
  \item \textbf{Unified dual-mode model:} the same model runs in streaming and
  full-context modes, and recovers most of the full-context quality in the latter.
\end{itemize}

\subsection{Challenges}
\label{sec:challenges}

The key issue in streamizing a model is \emph{context bounding}: how much 
context is appropriate (sufficient but not excessive) to decode the right content? 
This needs a target for what the model should emit
at time $t$, and a practical way to label that target at scale.

\textbf{Targets.} A forced aligner such as Qwen3-ForcedAligner
\citep{shi2026qwen3asr} gives word timestamps, which could serve as context
boundaries. However, a token's own acoustic features are often insufficient to determine it, and
later audio disambiguates; requiring emission at the timestamp would force the
model to guess.

\textbf{Residual interference.} More context is not always better. Revision of
earlier words as audio arrives is a known effect
\citep{mcgraw2012wordstability,shangguan2020stability}. In our setting, when we
feed audio progressively to a full-context model, the number of correct tokens
does not grow monotonically, because the newly appended tail disturbs earlier
tokens, a phenomenon we call \textit{residual interference}. A streaming
model must be trained to be resilient against such interference. 

Prior work bounds context by construction, through chunk-wise incremental
decoding \citep{shi2026qwen3asr} or limited look-ahead
\citep{wang2020transformerhybrid,moritz2020streamingtransformer}, rather than
learning when the available context suffices.

\section{Approach}

\subsection{Design Considerations}
Accuracy and latency conflict, and the stable-prefix property is trivial to
obtain if either is ignored.  The open problem is stability
\emph{together with} full-context accuracy and low latency.

We therefore adapt a full-context model with a \emph{frozen encoder}: no
causalization or attention masks. A pre-trained model brings strong
representations, data efficiency and a within-model reference for measuring
streaming loss \citep{kumar2025xlsrtransducer}. Its main liability is the
train--test mismatch: the encoder was trained on complete utterances, so its
representations of truncated audio differ from those of the full utterance.
XLSR-Transducer \citep{kumar2025xlsrtransducer}, Two-Pass Whisper
\citep{zhou2025twopasswhisper} and Chunked AED
\citep{zeineldeen2024chunkedaed} address the mismatch by training the encoder
with chunking or masking. Since our encoder is frozen, we address it in the
decoder, through longest-stable-prefix learning: the decoder learns to emit
\waittoken{} when the audio received so far does not determine the next token.
The mismatch is thus handled by deciding when to commit, not by changing the
representations.

Freezing the encoder keeps the bidirectional representations that give the
pre-trained model its accuracy, and leaves full-context decoding to depend only
on the decoder. Causalization is costly: 
\citet{krichli2026whisperrt} report that causalizing Whisper's encoder by LoRA
fine-tuning is worse than leaving it intact at chunk sizes above roughly
300\,ms. A primary motivation for causal encoders is computational efficiency, since a
bidirectional encoder must be re-run on the growing audio at each query. We
 show that this cost can be reduced with inference optimizations such as cache reuse and 
 \waittoken{}-terminated decoding (Section~\ref{sec:streaming-inference}).

The \waittoken{} token separates the timing decision (when to emit) from the
content decision (what to emit) in the supervision, rather than building timing
into the encoder. DSM and F-DSM \citep{takagi2026fdsm} also separate the two
decisions in LLM-based streaming ASR. Unlike F-DSM's factorized head, our \waittoken{} is a token in
the same vocabulary as the text tokens without any architectural change. This 
remains a \textit{recognition} task, and streaming behavior emerges when the model learns to emit the right content 
against residual interference, making near-lossless streaming conversion a natural
consequence of the design rather than an empirical coincidence.

\subsection{Architecture}

R2T2 is post-trained from Qwen3-ASR-1.7B, which consists of the Qwen3-1.7B LM, a projector, and an AuT encoder with 300M parameters and a hidden size of 1,024. AuT
is an attention-encoder-decoder (AED) based ASR model that downsamples Fbank features by a factor of eight, yielding a 12.5 Hz token rate. It also uses a dynamic flash attention window size ranging from 1 s to 8 s, which we utilize to implement encoder caching (Section \ref{sec:streaming-inference}). Other than that, there is no architectural change. In particular, no additional layers or heads are added.

\subsection{Longest Stable Prefix Learning}
\label{sec:lsp_learning_main}

We address the challenges mentioned in Section \ref{sec:challenges} with Longest Stable Prefix (LSP) learning. The term was first introduced by \citet{mcgraw2012wordstability} to describe an ideal path that an incremental speech recognizer's results could follow. Another related but different object was introduced for training in simultaneous translation by \citet{hoang2026dynamic}. In this section, we first present our definition and highlight their differences and the consequences.  We give the topic a detailed and rigorous treatment in Appendix \ref{sec:lsp-learning}.
\begin{definition}[Model-admissible final-anchored prefix]
\label{def:lsp}
The \emph{longest stable prefix} at time $t$ is
$L_t=\LSP(\mathcal{M},x,t)=\bar y_{1..m_t}$, where
\begin{equation}
  m_t \;=\; \max_{j \le t}\; n_j \cdot \mathbf{1}\!\left[\, y_j \prefeq \bar y \,\right],
  \label{eq:mt}
\end{equation}
and $m_t=0$ (so $L_t=\emptyset$) if no output so far is a prefix of $\bar y$.
\end{definition}

We call a label \emph{model-admissible} at $t$ if it \emph{equals} an output the
model produced from audio available at $t$, that is, $y_j$ for some $j\le t$; a
truncation of one output, or a string assembled from several, is not
model-admissible. In words, $L_t$ is the longest earlier output that is
\emph{entirely} a prefix of $\bar y$. It is therefore model-admissible, it never
shrinks, and $L_T=\bar y$.

\paragraph{Comparison.}
None of the three definitions says anything about accuracy, since $\bar y$ is itself a model
output. They differ along two axes (Table~\ref{tab:lsp-compare}): what the prefix
is anchored to, and whether the label is model-admissible.

The two anchors are close but different. Because $y_T=\mathcal{M}(x)=\bar y$,
the full-context output \emph{is} the last output of the run, so the all-future
criterion also requires agreement with $\bar y$; no definition is
$\bar y$-blind. What the all-future criterion adds is agreement with every
\emph{intermediate} output, and that is what can shorten it: a prefix shared by
$y_t$ and $y_T$ may be broken by some $y_{t'}$ with $t<t'<T$ and restored later.
The other two test the endpoint only.

\begin{example}[Appendix~\ref{sec:lsp-learning}, Example~\ref{ex:threeway}]
\label{ex:main-threeway}
Let $\bar y=$``a b c d e'' and let the run produce $y_1=$``a b c x'',
$y_2=$``a b'', $y_3=$``a b c d e''. At $t=2$ the all-future label and ours are
both ``a b''; the final-anchored label is ``a b c'', through the agreeing head of
$y_1$, whose tail diverges. No output of the run equals ``a b c''. At $t=3$ all
three name the whole utterance.
\end{example}

\begin{example}[The all-future label (Definition~\ref{def:allfuture}) is the strictest mid-run]
\label{ex:main-allfuture}
Let $\bar y=$``a b d'' and let the run produce $y_t=$``a b'', $y_{t+1}=$``a c'',
$y_{t+2}=$``a b d''. At $t$ the all-future label is ``a'', since its second token
does not survive $y_{t+1}$, while ours and the final-anchored label are ``a b''.
\end{example}

The final-anchored label contains the other two and is therefore the longest of
the three, but it is not model-admissible: Example~\ref{ex:main-threeway}
separates it from ours, Example~\ref{ex:main-allfuture} separates the all-future
label from both, and all three coincide once the raw outputs already form an
append-only chain (Propositions~\ref{prop:contain}--\ref{prop:coincide}).

\begin{table}[t]
\centering
\footnotesize
\setlength{\tabcolsep}{4pt}
\begin{tabular}{@{}llcl@{}}
\toprule
Formulation & Anchored to & Equals a whole earlier output & Credit \\
\midrule
All-future (Def.~\ref{def:allfuture})
  & all later outputs, $\bar y$ included
  & not in general & partial \\
Final-anchored (Def.~\ref{def:finalanchored})
  & $\bar y$
  & not in general
  & partial, $\max_{j\le t}\ell_j$ \\
Ours (Def.~\ref{def:lsp})
  & $\bar y$ & yes, by construction & all-or-nothing \\
\bottomrule
\end{tabular}
\caption{Three hindsight prefix definitions, with $\ell_j=|\lcp(y_j,\bar y)|$.
All-future is due to \citet{mcgraw2012wordstability}, final-anchored to
\citet{hoang2026dynamic}. The third column is model admissibility in the strict
sense; under the weak reading (prefix of an earlier output), all three would
qualify and the column would not discriminate.}
\label{tab:lsp-compare}
\end{table}

\paragraph{As a learning target.}
The final-anchored label is a partial-credit maximum, so it can name a string no
output equals: the agreeing head of an output whose tail the model itself
revised. Supervising on it credits a state in which the model had not settled, and
the amount credited is decided by comparing every past output with the complete
$\bar y$, hence by audio that has not arrived. Ours credits an output only if all
of it agrees with $\bar y$; the label is then always an output the model actually
produced (Proposition~\ref{prop:admissible}) and is the longest such prefix of
$\bar y$ (Theorem~\ref{thm:maximal}). The all-future label is a truncation of
$y_t$ and, as Example~\ref{ex:main-allfuture} shows, can be shortened by an
intermediate output that departs and returns; in general it equals no earlier
output.

The price is conservativeness. One diverging token zeroes the credit for that
step, so our label is never longer than the final-anchored one and biases the
model toward later emission. We accept this to keep admissibility, and measure
its effect in Section~\ref{sec:streamability}.

\section{Training Data Construction}
\label{sec:data-construction}

The training data for streamization comprise approximately
$1{,}500$ hours of proprietary Chinese and English audio.
ASR models can hallucinate on non-speech audio, degrade under
noise and reverberation, or translate instead of transcribing
when language cues conflict. Therefore we augment training examples to target these failure modes.
Noise augmentation includes pure-noise replacement and
mixed-noise injection, with additive-noise signal-to-noise
ratios (SNRs) sampled from 10--35~dB.
We also apply reverberation using real or simulated room
impulse responses. Additionally, we flip the language tag
for a subset of Chinese and English samples while keeping
the transcription target unchanged. 
%Detailed corpus and augmentation settings are listed in Appendix~\ref{app:configurations}.

\subsection{Streamability: Selecting the Right LSP Supervision}
\label{sec:streamability}

LSP is model-specific and its quality is actually measurable. Indeed, to select a good teacher or backbone model, it is necessary to quantify the model's potential for streamization. For this purpose we introduce the notion of \textit{streamability}. The length $l_{x,t} = |\LSP(\mathcal{M},\, x,\, t)|$ reflects the number of tokens $\mathcal{M}$ produces when seeing $x_{\leq t}$ that are in agreement with the output when it sees the full input sequence. From a streaming perspective, greater $l$ means shorter latency. If we marginalize over data, we can have for any time $t$

\begin{equation}
\label{eqn:streamability_unnormalized}
    l(\mathcal{M}, t) = \frac{1}{|\mathcal{N}(t)|}\sum_{x \in \mathcal{N}(t)}  |\LSP(\mathcal{M},\, x,\, t)|
\end{equation}

where $\mathcal{N}(t) = \{x : |x| \geq t\} $ is the set of data samples whose lengths are at least $t$, so it is meaningful to count their LSPs.  

We can plot the curve of $l(\mathcal{M}, t)$ against $t$. The two variables can also be normalized to make the patterns more prominent: each data point $(t, l_{x,t})$ is transformed into $(\tau_{x,t}, \bar{l}_{x,t})$ where $\tau_{x,t} =  t/|x|$, and $\bar{l}_{x,t} = |\LSP(\mathcal{M}, x, t)|/|\mathcal{M}(x)|$ with $\mathcal{M}(x)$ being the full-utterance transcript. $(\tau_{x,t}, \bar{l}_{x,t})$ is then aggregated across $x$:

\begin{equation}
\label{eqn:streamability_normalized}
    \bar{l}(\mathcal{M}, \tau) = \frac{1}{|\mathcal{N}(0)|}\sum_{x} \bar{l}_{x,t=\tau|x|}
\end{equation}

We can also further aggregate across time and obtain $s(\mathcal{M})$ = AUC (area under the curve), which we define as $\mathcal{M}$'s \textit{streamability}. It is a single measurement for a model's responsiveness when fed with an incremental audio sequence. We use it to evaluate both a base model's streamization potential and a teacher model's annotation quality. 

\begin{figure}[!htbp]
\centering
\input{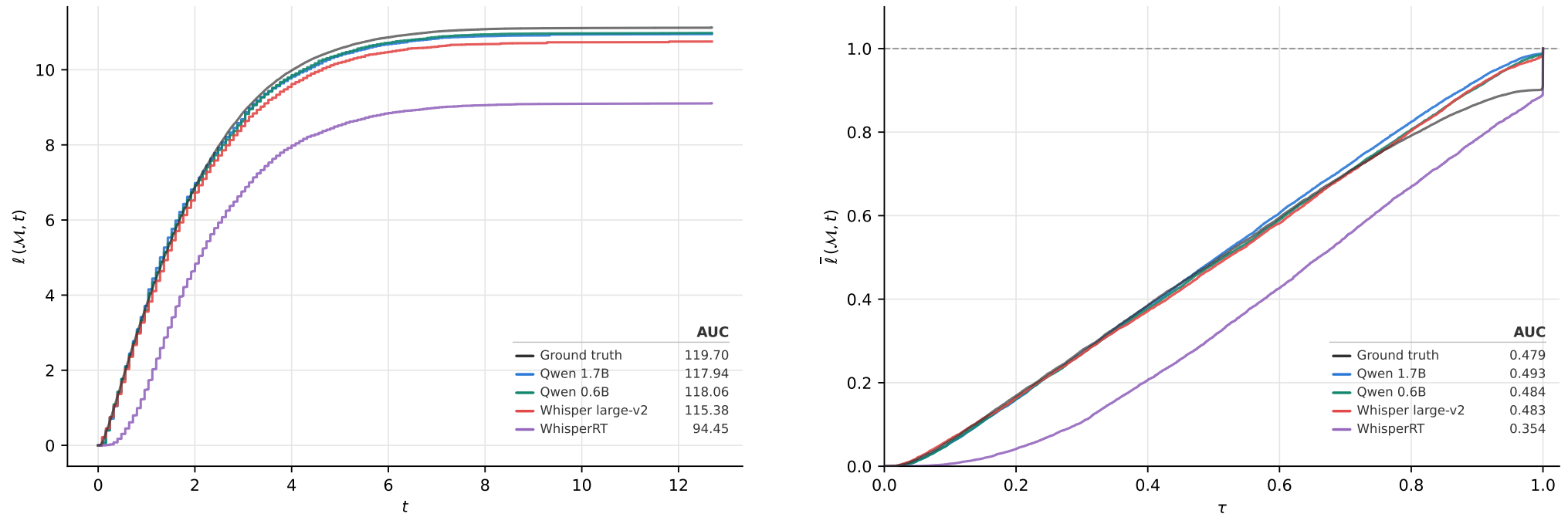}
\caption{Streamability plots on
1,860 Buckeye utterances. The curves compare Qwen3-ASR-1.7B,
Qwen3-ASR-0.6B, Whisper large-v2, and WhisperRT, with manually
timestamped ground truth as a timing reference. Left: unnormalized (Equation \ref{eqn:streamability_unnormalized}). 
Right: normalized (Equation \ref{eqn:streamability_normalized}).
Insets report AUC over the displayed domains.}
\label{fig:lsp-teacher-comparison}
\end{figure}

Figure~\ref{fig:lsp-teacher-comparison} shows both the unnormalized (Equation \ref{eqn:streamability_unnormalized}, left) and normalized (Equation \ref{eqn:streamability_normalized}, right) streamability plots for a number of models on 1,860 English utterances from the selected Buckeye
Speech Corpus \citep{pitt2007buckeye} subset, which contains manually verified
word-level timestamps providing a timing reference. For each model
and utterance, we construct LSP targets from cumulative audio prefixes
at 80\,ms intervals. 

Qwen3-ASR-1.7B achieves the highest normalized streamability of 0.493,
indicating the greatest potential for streaming adaptation among the evaluated models. WhisperRT, which is a streaming variant of Whisper large-v2 constructed by fine-tuning its encoder with a causal attention mask and adapting the encoder-decoder alignment, performs the worst among the models. Interestingly, its streamability is worse than its full-context base model's (red line). This is caused by the instability of its output, resulting in much shorter LSPs. 

Several curves exceed the timestamp-based reference (black line) during the
final part of normalized time. This indicates greater relative progress
toward their respective final transcripts. Because the model and
ground-truth transcripts can differ in content and length, this
observation alone does not establish that corresponding words were
correctly predicted before their acoustic ends. Such an interpretation
would require an additional word-aligned analysis.

%Note that under our definition of LSP, which is anchored to the model's own output from the full utterance, streamability says nothing %about recognition quality, which must be evaluated separately using metrics such as WER. This is also a very important metric when %selecting a model. This is why we do not choose Qwen3-ASR-0.6B even though it has a slightly higher raw streamability in the left %panel, 

We find in practice that $s(\mathcal{M})$ is a good indicator of a full-context model's responsiveness. We use it to assess the model's potential for streamization. However, whether it can serve as a solid gauge for general streamization (e.g., simultaneous translation) remains open. In addition, \method{} uses a self-distillation approach, avoiding cross-model teacher--student distribution mismatch. Whether a stronger teacher model could produce more capable streaming models is also open. We believe both issues are interesting directions for future exploration.

\subsection{LSP Annotation}
\label{sec:lsp-supervision}

Given a model $\mathcal{M}$, the audio sequence $x=(x_1, \ldots, x_T)$ is fed into $\mathcal{M}$ progressively frame-by-frame at an interval $\delta=80\,\mathrm{ms}$ to produce $\LSP(\mathcal{M},\, x,\, t)$. During the same process, we also record the intervals where the model's output agrees with its LSP.
%, i.e.,
%
%\begin{displaymath}
%     I_t \leftarrow  I_t \cup t  \,\, \mathrm{if} \,\, \mathcal{M}(x_{\leq t}) = \LSP(\mathcal{M}, x, t) 
%\end{displaymath}
%
These are the ``stable'' regions of the model. We merge consecutive stable regions into intervals and label them by their starting and ending timestamps $I_i = [s_i, e_i]$. Each interval is a maximal run of consecutive audio-prefix endpoints at which the teacher hypothesis equals the LSP and that prefix remains unchanged. The start $s_i$ is the first endpoint in the run, and the end $e_i$ is the last; an uncertain continuation or a change of prefix ends the run. We store each interval with its corresponding stable prefix. 

Outside the stable intervals, the model's outputs are unstable. Therefore, the model should learn when to defer uncertain continuations with a special \waittoken{} token. 

During training, we cut each sample at a random position $t$ and discard the subsequence $x_{>t}$, simulating the situation where the model sees only $x_{\leq t}$. The corresponding label is 

 \begin{equation}
    y_t^*
    =
    \begin{cases}
        \LSP(\mathcal{M}, x, t),
        & \exists i:\ s_i\leq t \leq e_i,\\[2pt]
        \LSP(\mathcal{M}, x, t)\oplus\waittoken{},
        & \text{otherwise}.
    \end{cases}
    \label{eq:lsp-target}
\end{equation}

\begin{figure}[!htbp]
\centering
\begingroup

% Colors
\definecolor{figBlue}{HTML}{287C91}
\definecolor{figAmber}{HTML}{B78339}
\definecolor{figInk}{HTML}{29343D}

\begin{tikzpicture}[
    x=1cm,
    y=1cm,
    font=\scriptsize,
    stable/.style={
        fill=figBlue!10,
        draw=figBlue!45,
        text=figBlue!85!black,
        rounded corners=1pt,
        line width=0.3pt,
        inner xsep=3pt,
        inner ysep=1pt
    },
    unstable/.style={
        fill=figAmber!10,
        draw=figAmber!45,
        text=figAmber!85!black,
        rounded corners=1pt,
        line width=0.3pt,
        inner xsep=3pt,
        inner ysep=1pt
    }
]

% ------------------------------------------------------------
% Layout parameters
% ------------------------------------------------------------

\def\yfirst{-0.48}
\def\ystep{0.34}

% Token positions:
% I = 2.10
% have = 3.20
% to/too/two = 4.45
% cats. = 5.75

% ------------------------------------------------------------
% Helper commands
% ------------------------------------------------------------

% Place a token at audio-prefix endpoint #1
\newcommand{\ASRunit}[3]{%
    \pgfmathsetmacro{\ry}{
        \yfirst-(#1-1)*\ystep
    }
    \node[text=figInk!80]
        at (#2,\ry) {\texttt{#3}};
}

% Highlight the final predicted token
\newcommand{\ASRfinal}[4]{%
    \pgfmathsetmacro{\ry}{
        \yfirst-(#1-1)*\ystep
    }
    \node[#4]
        at (#2,\ry) {\texttt{#3}};
}

% Draw an extracted interval
% #1 = first audio-prefix endpoint
% #2 = last audio-prefix endpoint
% #3 = agreement-interval index
% #4 = start time
% #5 = end time
\newcommand{\ASRinterval}[5]{%
    \pgfmathsetmacro{\ya}{
        \yfirst-(#1-1)*\ystep
    }
    \pgfmathsetmacro{\yb}{
        \yfirst-(#2-1)*\ystep
    }
    \pgfmathsetmacro{\ym}{
        (\ya+\yb)/2
    }

    \draw[
        figBlue,
        line width=2pt,
        line cap=round
    ]
        (7.35,\ya) -- (7.35,\yb);

    \draw[figBlue, line width=0.6pt]
        (7.35,\ya) -- (7.50,\ya);

    \draw[figBlue, line width=0.6pt]
        (7.35,\yb) -- (7.50,\yb);

    \node[
        anchor=west,
        text=figBlue!85!black
    ] at (7.65,\ym)
        {$I_{#3}=[#4,#5]$};
}

% ------------------------------------------------------------
% Header
% ------------------------------------------------------------

\node[
    anchor=west,
    text=figInk
] at (0,0.80)
{
    Reference: \texttt{I have two cats.}
};

\node[
    anchor=east,
    font=\scriptsize\bfseries
] at (0.85,0.15)
    {$t_k$ (s)};

\node[
    font=\scriptsize\bfseries
] at (3.80,0.15)
    {Hypothesis};

\node[
    font=\scriptsize\bfseries
] at (6.70,0.15)
    {Status};

\node[
    anchor=west,
    font=\scriptsize\bfseries
] at (7.65,0.15)
    {Intervals};

\draw[gray!45, line width=0.5pt]
    (0,0) -- (10.6,0);

% ------------------------------------------------------------
% Teacher audio-prefix endpoints
% ------------------------------------------------------------

\foreach \k/\t in {
    1/0.08,
    2/0.16,
    3/0.24,
    4/0.32,
    5/0.40,
    6/0.48,
    7/0.56,
    8/0.64,
    9/0.72,
    10/0.80,
    11/0.88,
    12/0.96,
    13/1.04,
    14/1.12
}{
    \pgfmathsetmacro{\ry}{
        \yfirst-(\k-1)*\ystep
    }

    \node[
        anchor=east,
        text=figInk
    ] at (0.85,\ry) {\t};

    \draw[gray!13]
        (1.25,\ry-0.17) --
        (7.00,\ry-0.17);

    \fill[gray!55]
        (1.15,\ry) circle (1pt);
}

% Timeline
\draw[
    gray!35,
    line width=0.5pt
]
    (1.15,-0.48) -- (1.15,-4.90);

% Redraw audio-prefix endpoints above the line
\foreach \k in {1,...,14}
{
    \pgfmathsetmacro{\ry}{
        \yfirst-(\k-1)*\ystep
    }

    \fill[gray!55]
        (1.15,\ry) circle (1pt);
}

% ------------------------------------------------------------
% Full ASR transcriptions
% ------------------------------------------------------------

% "I"
\foreach \k in {1,...,14}
{
    \ASRunit{\k}{2.10}{I}
}

% "have"
\foreach \k in {3,...,14}
{
    \ASRunit{\k}{3.20}{have}
}

% "to" at 0.48 and 0.56
\foreach \k in {6,7}
{
    \ASRunit{\k}{4.45}{to}
}

% "too" at 0.64 and 0.72
\foreach \k in {8,9}
{
    \ASRunit{\k}{4.45}{too}
}

% "two" from 0.80 onward
\foreach \k in {10,...,14}
{
    \ASRunit{\k}{4.45}{two}
}

% "cats." from 0.96 onward
\foreach \k in {12,...,14}
{
    \ASRunit{\k}{5.75}{cats.}
}

% ------------------------------------------------------------
% Stable and unstable terminal predictions
% ------------------------------------------------------------

% Stable "I"
\foreach \k in {1,2}
{
    \ASRfinal{\k}{2.10}{I}{stable}
}

% Stable "have"
\foreach \k in {3,4,5}
{
    \ASRfinal{\k}{3.20}{have}{stable}
}

% Unstable "to"
\foreach \k in {6,7}
{
    \ASRfinal{\k}{4.45}{to}{unstable}
}

% Unstable "too"
\foreach \k in {8,9}
{
    \ASRfinal{\k}{4.45}{too}{unstable}
}

% Stable "two"
\foreach \k in {10,11}
{
    \ASRfinal{\k}{4.45}{two}{stable}
}

% Stable "cats."
\foreach \k in {12,...,14}
{
    \ASRfinal{\k}{5.75}{cats.}{stable}
}

% ------------------------------------------------------------
% Stability indicators
% ------------------------------------------------------------

% Agreement endpoints
\foreach \k in {
    1,2,3,4,5,10,11,12,13,14
}
{
    \pgfmathsetmacro{\ry}{
        \yfirst-(\k-1)*\ystep
    }

    \fill[figBlue]
        (6.70,\ry) circle (1.6pt);
}

% Endpoints requiring WAIT
\foreach \k in {6,7,8,9}
{
    \pgfmathsetmacro{\ry}{
        \yfirst-(\k-1)*\ystep
    }

    \fill[figAmber]
        (6.70,\ry) circle (1.6pt);
}

% ------------------------------------------------------------
% Extracted stable-prefix intervals
% ------------------------------------------------------------

\ASRinterval{1}{2}{1}{0.08}{0.16}

\ASRinterval{3}{5}{2}{0.24}{0.40}

\ASRinterval{10}{11}{3}{0.80}{0.88}

\ASRinterval{12}{14}{4}{0.96}{1.12}

% ------------------------------------------------------------
% Legend
% ------------------------------------------------------------

\draw[gray!35, line width=0.4pt]
    (0,-5.15) -- (10.6,-5.15);

\fill[figBlue]
    (1.15,-5.47) circle (1.6pt);

% Blue indicator
\node[
    anchor=west,
    text=figInk
] at (1.35,-5.47)
    {Stable};

\fill[figAmber]
    (4.05,-5.47) circle (1.6pt);

% Orange indicator
\node[
    anchor=west,
    text=figInk
] at (4.25,-5.47)
    {\waittoken};

\end{tikzpicture}

\endgroup
\caption{\textbf{Stable-prefix interval extraction.}
A teacher ASR model decodes progressively longer audio
prefixes every 80\,ms. Each row shows the complete
recognition hypothesis. Blue indicates predictions
matching the stable reference prefix, whereas orange
indicates uncertain continuations requiring waiting.
Intervals $I_i=[s_i,e_i]$ describe the persistence
of each stable prefix. Checkpoints with 
unstable continuations fall outside these intervals.}
\label{fig:lsp_construction}

\end{figure}
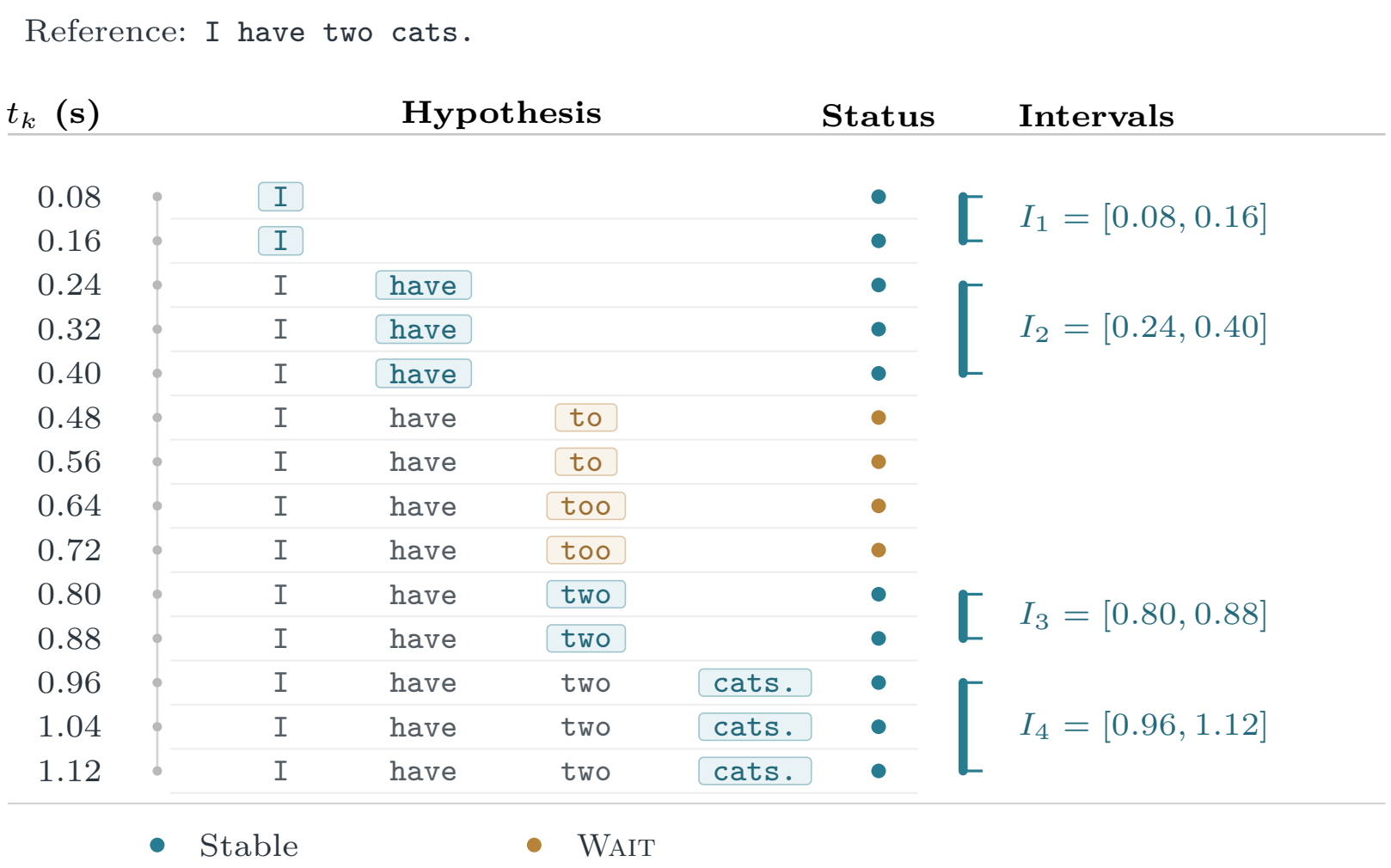

Figure~\ref{fig:lsp_construction} illustrates the construction:
blue marks agreement with a stable prefix, and orange marks
uncertain continuations for successive hypotheses.

The extracted stable-prefix intervals and waiting
annotations provide temporal supervision for streaming
commitment. Cuts within stable intervals supervise the stable
transcription, while cuts in the gaps add an explicit waiting
signal. Only the prefix content is inherited from the preceding
annotation endpoint; the waiting decision uses $t$ directly.
Before the first endpoint, the target consists only of
$\waittoken{}$. At $t=T$, we use the complete reference
transcription, preserving the original offline recognition
objective.

This formulation jointly supervises lexical recognition
and emission timing. By separating stable content from
uncertain continuations, the model learns to preserve
previously recognized text while deferring predictions
that require additional acoustic context, supporting
the output monotonicity constraint.

\section{Training}
\label{sec:training}

\sysname{}  is trained in two stages: supervised fine-tuning (SFT) and reinforcement learning (RL). SFT familiarizes the model with LSP output and the  \waittoken{} token and improves its resilience to residual tail interference common in streaming but not experienced during its pre-training. RL further enforces the model's ability to trade off quality and speed while suppressing undesired behaviors such as unsupported continuations, repetitions, language inconsistency, and hallucinated speech on no-speech inputs. The audio encoder and projector are frozen during both stages. This preserves the pre-trained acoustic representations and their projection into the language decoder's input space. Freezing these components also reduces the number of trainable parameters and avoids storing their gradients and optimizer states.

\subsection{Supervised Fine-Tuning}
\label{sec:training-recipe}
\label{sec:sft}

SFT uses a cross-entropy loss to adapt the model to LSP targets
and the \waittoken{} action. Training runs for one epoch
using fused AdamW, an initial learning rate of $10^{-5}$,
$2\%$ warmup followed by linear decay, and an effective
global batch size of 64. During SFT, only the language-decoder
parameters are tuned. This focuses adaptation on the LSP response format and emission
timing in the decoder. 

\paragraph{Acoustic Context and Prefix Sampling.}
The encoder attention window is also sampled between $1$
and $8$ seconds to train commitment behavior under varying
amounts of acoustic context, while preserving bidirectional
attention within each window.

Each example uses the complete utterance and its full
transcription with probability $10\%$; otherwise, we sample
a prefix cut ($90\%$). Conditional on choosing a prefix cut,
the supervision timestamp is sampled within an LSP stable
interval with probability $50\%$ and in a gap between these
intervals otherwise. Cuts may fall between annotation endpoints.
The recognizer receives audio from the start of the utterance
to the sampled timestamp.
Complete-utterance examples supervise offline recognition;
interval cuts supervise stable-prefix transcription, and gap
cuts additionally supervise waiting through
Equation~\ref{eq:lsp-target}.

\subsection{Reinforcement Learning}
\label{sec:rl-training}
Starting from the SFT model, we apply Group Relative
Policy Optimization (GRPO)~\citep{shao2024deepseekmath} to
refine recognition quality and waiting behavior. For each
prompt, the model samples multiple streaming trajectories
and uses their relative rewards to guide policy optimization.
Rollouts span one, two, or four streaming turns, allowing
continuations to be assessed as additional audio arrives.
\paragraph{Reward Design.}
\label{sec:rl-reward}

To refine the model's recognition accuracy and emission
behavior, we employ a multi-dimensional reward function
during GRPO training. Character Error Rate (CER) provides
the primary recognition signal, while complementary
rewards penalize redundant predictions, repetitions,
language inconsistencies, and premature continuation
at uncertain acoustic boundaries.

The reward consists of eight components, summarized in
Table~\ref{tab:reward}. The Tail and Cut Marker components
are designed to discourage unsupported continuation and
supervise the placement of the \waittoken{} token.
The remaining components score recognition fidelity and
penalize selected hallucination patterns. Component-specific
effects require controlled ablations.

For a generated response $y_t$ and its supervision target
$y_t^{\ast}$, the raw reward combines the individual
components using predefined weights, with special-case
rules for empty, exact-match, and other degenerate
predictions. The final reward is obtained through
a smooth nonlinear transformation:

\begin{equation}
    R(y_t,y_t^{\ast})
    =
    \tanh\!\left(
        3R_{\mathrm{raw}}(y_t,y_t^{\ast})
    \right),
    \label{eq:final-reward}
\end{equation}

where $R_{\mathrm{raw}}$ denotes the weighted reward
after applying the special-case rules.

Details for both stages are provided in Appendix~\ref{app:configurations}.

% Streaming inference section (Section 6 of the R2T2 technical report).
% Preamble requirements (in addition to the paper's existing packages):
%   \usepackage{tikz}
%   \usetikzlibrary{arrows.meta,decorations.pathreplacing,patterns}
% Bibliography: merge inference_refs.bib into the paper's bibliography.
% Labels this section expects from the rest of the paper (numbers as in
% "R2T2 (4).pdf"; inference_preview.tex defines stand-ins for them):
%   sec:related-stabilization  2.2  Incremental Hypothesis Stabilization
%   sec:lsp                    4.1  Longest Stable Prefix (LSP)
%   sec:two-stage              5.2  Two-Stage Training
%   sec:latency-computation               7.2  Latency Computation Method
%   sec:eval-results           7.3  Evaluation results
%   sec:sweep                  7.4  Chunk and revision-window sensitivity
%   tab:results-en             Table 3 (English WER)
%   tab:results-zh             Table 4 (Chinese CER)
% figures/inference_verification.tex (pre-fill verification schematic) is no
% longer used here; it can go in an appendix.
% Shared colors, TikZ styles, and draft macros for the streaming-inference
% section and its figures. Requires: xcolor, tikz, and TikZ libraries
% arrows.meta, decorations.pathreplacing, patterns (plus pgfplots, compat 1.18,
% for the optional figures/inference_prefill_growth.tex).
% Palette matches figures/revision_aware_decoding.tex.

\definecolor{rdInk}{HTML}{253247}
\definecolor{rdMuted}{HTML}{758194}
\definecolor{rdLine}{HTML}{C7CFD9}
\definecolor{rdGrayLight}{HTML}{F5F7F9}
\definecolor{rdBlue}{HTML}{397CAD}
\definecolor{rdBlueLight}{HTML}{EDF4FA}
\definecolor{rdAmber}{HTML}{B7791F}
\definecolor{rdAmberLight}{HTML}{FCF3E3}
\definecolor{rdTeal}{HTML}{247A67}
\definecolor{rdTealLight}{HTML}{EAF5EF}
\definecolor{rdRed}{HTML}{BE493F}
\definecolor{rdRedLight}{HTML}{FCEFEB}

\tikzset{
  inf/base/.style={font=\sffamily\small, text=rdInk, line width=0.6pt},
  inf/blk/.style={rounded corners=1.5pt, line width=0.6pt},
  inf/reused/.style={inf/blk, draw=rdBlue!70, fill=rdBlueLight},
  inf/recomp/.style={inf/blk, draw=rdAmber!80, fill=rdAmberLight},
  inf/decoded/.style={inf/blk, draw=rdInk!70, densely dashed, fill=white},
  inf/evicted/.style={inf/blk, draw=rdLine, fill=rdGrayLight,
    postaction={pattern=north east lines, pattern color=rdLine}},
  inf/future/.style={inf/blk, draw=rdLine, densely dotted},
  inf/neutral/.style={inf/blk, draw=rdLine, fill=rdGrayLight},
  inf/match/.style={inf/blk, draw=rdTeal!70, fill=rdTealLight},
  inf/reject/.style={inf/blk, draw=rdRed, fill=rdRedLight, line width=0.9pt},
  inf/sub/.style={font=\sffamily\scriptsize, text=rdMuted},
  inf/note/.style={font=\sffamily\footnotesize, text=rdMuted},
  inf/brace/.style={decorate, decoration={brace, amplitude=4pt, mirror},
    line width=0.6pt},
  inf/arrow/.style={-{Stealth[length=1.8mm, width=1.3mm]}, line width=0.7pt},
}

% Draft placeholders: author-confirmed facts and unmeasured table cells.
\providecommand{\tbd}[1]{\textcolor{rdRed}{[#1]}}
\providecommand{\ph}{\textcolor{rdMuted}{xx.x}}

\section{Streaming Inference}
\label{sec:streaming-inference}
\suppressfloats[t]% keep Table 2 from floating above this heading

The encoder produces one audio token every $\delta = 80$ ms but the model is configurable to 
decode at a coarse granularity, i.e., performing chunk-wise decoding with chunk size $\Delta = k\delta$\,
 to trade quality for speed.  At the $\Delta=160$ ms chunk used in our main results, \method{} performs 6.25 updates
per second for each stream. A less-than-1 RTF (Real Time Factor) requires that every update finishes 
before the next chunk arrives. This section describes how we keep the cost of an update low enough to meet that
assumption.

For models whose decoder input places audio before the transcript,
repeated-prefix decoding reprocesses accumulated audio and text, as in
Qwen3-ASR-1.7B streaming \citep{shi2026qwen3asr}. In the 2\,s, $U=5$
reference in Table~\ref{tab:recognition-results}, each update re-encodes
all audio received so far and pre-fills the decoder with the audio
representations and the previous transcript, excluding its last five
tokens, which are regenerated. The first two chunks are decoded
without a transcript prefix, and no cache is retained across updates
\citep{qwen2026qwen3asrcode}.

To reduce cost without modifying the model or the output monotonicity constraint, we introduce
 four techniques: cache reuse (Section~\ref{sec:cache-reuse}), \waittoken{}-terminated decoding (Section~\ref{sec:wait-terminated-decoding}), 
 bounded context (Section~\ref{sec:bounded-context}), and pre-fill verification of the revisable suffix (Section~\ref{sec:prefill-verification}). 
 Through LSP training (Section \ref{sec:training}), the model
emits only the stable prefix supported by the received audio and defers the
uncertain tail with \waittoken{}. Its output therefore never needs to be rolled
back, and an update that predicts \waittoken{} ends right after the pre-fill.
Recomputation of old audio is removed by caching, which the 8-s attention
windows of the \qwen{} encoder make possible, and the context is bounded so
that the cost per update does not grow over long streams.  None of these optimizations changes the trained model.

At inference, a provisional-token budget $U$ can be used to delay the
commitment of a suffix of predicted text so that subsequent
updates can reconsider it. This withholding principle is
related to the hold-$n$ policy of
\citet{liu2020partialhypothesis}. For $U>0$, the decoder may use the provisional suffix,
but only released text enters the committed history. % $y_t$.

Table~\ref{tab:inf-work} compares the resulting work per update with that of
\qwen{} and of a causal interleaved design~\citep{seide2024speechreallm}, which
we include as a theoretical reference and did not implement. In that design,
the encoder is fully causal, and the decoder input alternates between audio
chunks and the text committed after each of them. Since no earlier state ever
changes, each update processes only the new chunk. We did not adopt this
design for two reasons. First, it requires training a causal encoder and
retraining the decoder on interleaved sequences, whereas \method{} keeps the
pre-trained \qwen{} encoder frozen. Second, it cannot
revise its output, because generated text is never re-read under later audio;
this rules out the quality-priority mode ($U=1$).
The price of keeping the \qwen{} architecture is that \method{} re-encodes the
active window and re-pre-fills the retained text at every update. Both are
bounded, and the pre-fill processes them in parallel rather than serially.

\begin{table}[h]
\centering\small
\begin{tabular}{@{}lccccc@{}}
\toprule
Design & Encoder & Decoder pre-fill & Serial steps & Revisable & Trained \\
\midrule
QS$^{\dagger}$ & all audio & all audio + text & $5\,+$ new & last 5 & --- \\
\method{} & active window & active window + text & new & last $U$
  & \shortstack{language\\decoder} \\
CI$^{\ddagger}$ & new chunk & new chunk & new & none
  & enc.\ + dec. \\
\bottomrule
\end{tabular}
\caption{Work per update for three streaming designs. All audio grows with
the stream at 12.5 positions per second, the active window holds at most 8\,s
(100 positions), and text is the retained transcript, which is bounded only in
\method{} (Section \ref{sec:bounded-context}). New is the number of newly generated
tokens, which is zero when \method{} predicts \waittoken{}. Revisable gives
the trailing tokens that the next update may change: \qwen{} regenerates them,
whereas \method{} verifies them in the pre-fill. Trained lists the components
trained on top of \qwen{}. For \method{}, only the language decoder is trained;
the audio encoder and projector remain frozen.
$^{\dagger}$ \qwen{} Streaming. $^{\ddagger}$ Causal interleaved (theoretical; not implemented). }
\label{tab:inf-work}
\end{table}

% ---------------------------------------------------------------------------
\subsection{Cache Reuse}
\label{sec:cache-reuse}

\begin{figure}[t]
\centering
\scalebox{.8}{% Decoder layout and cache regions at update k (N = 1 retained window).
% Input from a figure environment; requires figures/inference_style.tex.
\begin{tikzpicture}[inf/base]

% ---- Row labels -----------------------------------------------------------
\node[anchor=east, align=right] at (-0.2,3.6)
  {\footnotesize\bfseries Encoder\\[-1pt]{\scriptsize\mdseries\color{rdMuted}8-s windows}};
\node[anchor=east, align=right] at (-0.2,1.4)
  {\footnotesize\bfseries Decoder\\[-1pt]{\scriptsize\mdseries\color{rdMuted}update $k$}};

% ---- Time axis ------------------------------------------------------------
\draw[inf/arrow, draw=rdMuted] (0,4.35) -- (10.1,4.35)
  node[right, inf/note] {time};
\foreach \x/\t in {0/0, 3.2/8, 6.4/16, 9.6/24} {
  \draw[rdMuted] (\x,4.35) -- (\x,4.25);
  \node[above, inf/note, inner sep=1.5pt] at (\x,4.35) {\t\,s};
}
\draw[rdAmber, line width=0.9pt] (8.4,4.43) -- (8.4,4.2);
\node[above, font=\sffamily\footnotesize, text=rdAmber, inner sep=1.5pt]
  at (8.4,4.43) {$a_k$};

% ---- Encoder windows ------------------------------------------------------
\draw[inf/evicted] (0.03,3.2) rectangle (3.17,4.0);
\node[align=center, text=rdMuted] at (1.6,3.6)
  {\footnotesize window 1\\[-2pt]\scriptsize evicted};
\draw[inf/reused] (3.23,3.2) rectangle (6.37,4.0);
\node[align=center] at (4.8,3.6)
  {\footnotesize window 2\\[-2pt]{\scriptsize\color{rdBlue}completed, cached}};
\draw[inf/future] (8.4,3.2) rectangle (9.57,4.0);
\node[inf/sub, align=center] at (8.985,3.6) {not yet\\[-2pt]received};
\draw[inf/recomp] (6.43,3.2) rectangle (8.4,4.0);
\node[align=center] at (7.415,3.6)
  {\footnotesize window 3\\[-2pt]{\scriptsize\color{rdAmber}active}};

% ---- Encoder outputs projected into the decoder input --------------------
\fill[rdBlue, opacity=0.10]
  (3.23,3.2) -- (6.37,3.2) -- (4.42,1.8) -- (1.28,1.8) -- cycle;
\fill[rdAmber, opacity=0.14]
  (6.43,3.2) -- (8.4,3.2) -- (6.42,1.8) -- (4.48,1.8) -- cycle;

% ---- Decoder input sequence ----------------------------------------------
\draw[inf/reused] (0.03,1.0) rectangle (1.22,1.8);
\node[align=center] at (0.625,1.4) {\footnotesize prompt\\[-2pt]\scriptsize\color{rdMuted}head};

\draw[inf/reused] (1.28,1.0) rectangle (4.42,1.8);
\node[align=center] at (2.85,1.4)
  {\footnotesize window 2\\[-2pt]\scriptsize\color{rdMuted}$\approx$100 positions};

\draw[inf/recomp] (4.48,1.0) rectangle (6.42,1.8);
\node[align=center] at (5.45,1.4)
  {\footnotesize window 3\\[-2pt]\scriptsize\color{rdMuted}2--100 pos.};

\draw[inf/recomp] (6.48,1.0) rectangle (7.67,1.8);
\node[align=center] at (7.075,1.4) {\footnotesize prompt\\[-2pt]\scriptsize\color{rdMuted}tail};

\draw[inf/recomp] (7.73,1.0) rectangle (10.27,1.8);
\node[align=center] at (9.0,1.4)
  {\footnotesize committed $c_{k-1}$\\[-2pt]\scriptsize\color{rdMuted}fixed text};

\draw[inf/recomp] (10.33,1.0) rectangle (12.67,1.8);
\node[align=center] at (11.5,1.4)
  {\footnotesize revisable $r_{k-1}$\\[-2pt]\scriptsize\color{rdMuted}$\le U$, verified};

\draw[inf/decoded] (12.73,1.0) rectangle (14.27,1.8);
\node[align=center] at (13.5,1.4)
  {\footnotesize new text\\[-2pt]\scriptsize\color{rdMuted}or \texttt{WAIT}};

% Why retained text is prefilled again.
\draw[draw=rdMuted, line width=0.5pt] (7.73,1.93) -- (12.67,1.93);
\foreach \x in {7.73,12.67} \draw[draw=rdMuted, line width=0.5pt] (\x,1.93) -- (\x,1.86);
\node[inf/sub, anchor=south, inner sep=2pt] at (10.2,1.93)
  {fixed tokens, updated acoustic context};

% ---- Cost braces ----------------------------------------------------------
\draw[inf/brace, draw=rdBlue] (0.03,0.88) -- (4.42,0.88);
\node[font=\sffamily\footnotesize, text=rdBlue] at (2.225,0.5)
  {KV cache reused};
\draw[inf/brace, draw=rdAmber] (4.48,0.88) -- (12.67,0.88);
\node[font=\sffamily\footnotesize, text=rdAmber] at (8.575,0.5)
  {re-prefilled at every update};
\draw[inf/brace, draw=rdInk!70] (12.73,0.88) -- (14.27,0.88);
\node[font=\sffamily\footnotesize, align=center] at (13.38,0.42)
  {decoded\\[-2pt]{\scriptsize\color{rdMuted}skipped on \texttt{WAIT}}};

% ---- Legend ---------------------------------------------------------------
\draw[inf/reused]  (10.4,3.84) rectangle (10.8,4.06);
\node[anchor=west, font=\sffamily\footnotesize] at (10.87,3.95) {cached};
\draw[inf/recomp]  (10.4,3.46) rectangle (10.8,3.68);
\node[anchor=west, font=\sffamily\footnotesize] at (10.87,3.57) {recomputed per update};
\draw[inf/decoded] (10.4,3.08) rectangle (10.8,3.30);
\node[anchor=west, font=\sffamily\footnotesize] at (10.87,3.19) {decoded serially};
\draw[inf/evicted] (10.4,2.70) rectangle (10.8,2.92);
\node[anchor=west, font=\sffamily\footnotesize] at (10.87,2.81) {evicted};

\end{tikzpicture}}
\caption{Decoder input at update $k$ with $W=1$ retained window. The encoder
outputs and KV cache of completed windows are reused, whereas the active
window, the prompt tail, and all retained text are pre-filled again at every
update because new audio precedes the text.}
\label{fig:inf-layout}
\end{figure}
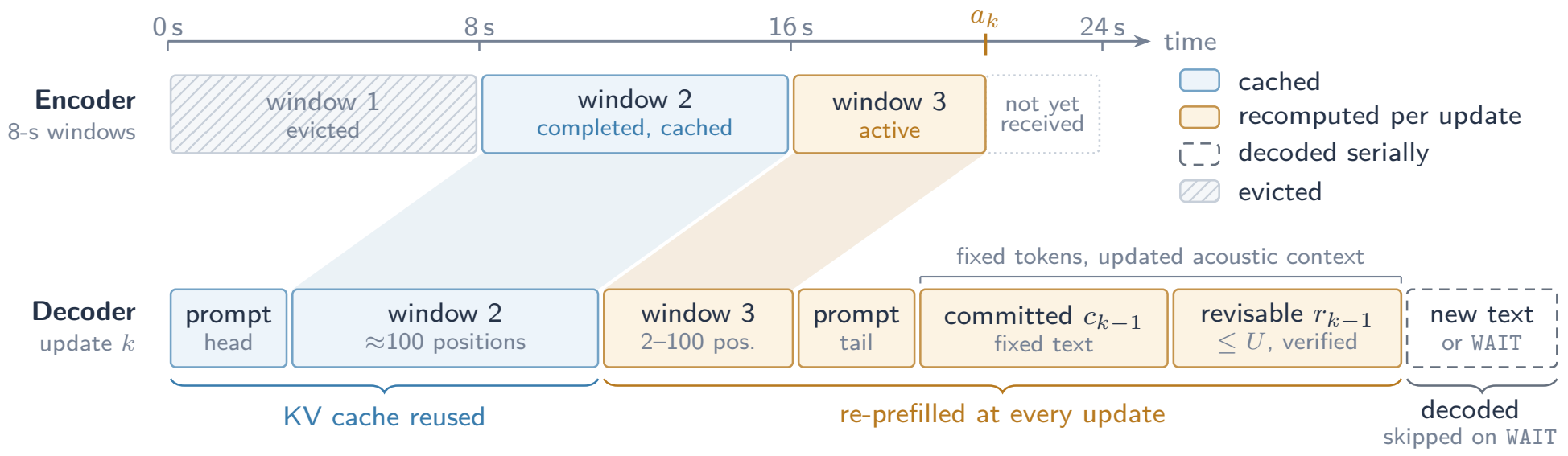

Figure~\ref{fig:inf-layout} shows the decoder input at update $k$: the prompt
head, the encoder outputs of the retained audio, the prompt tail, the committed
transcript $c_{k-1}$, and, if $U>0$, the revisable suffix $r_{k-1}$ of at most
$U$ tokens.

\textbf{Encoder-Side Caching.} The \qwen{} encoder attends within
non-overlapping windows, whose size we fix to 8\,s at inference. Once a window
is complete, its outputs no longer depend on later audio and can be cached.
Only the active window is re-encoded at each update; it grows from 2 to 100
positions over its 50 updates.

\textbf{Decoder-Side Caching.} The prompt head and the completed windows form
a prefix of the decoder input that is identical across updates, so their KV
cache is reused. Everything after this prefix is pre-filled again at each
update: the active window, the prompt tail, and the retained text. This
includes the committed text: its tokens are fixed, but the audio preceding it
has changed, and so have its KV entries.

% ---------------------------------------------------------------------------
\subsection{\texorpdfstring{\waittoken{}}{WAIT}-Terminated Decoding}
\label{sec:wait-terminated-decoding}

The prediction at the last pre-fill position is the first action of each
update. If it is \waittoken{}, the update ends without any serial decoding
step; otherwise, the model decodes greedily until it predicts \waittoken{} or
EOS. The number of serial steps therefore follows the amount of newly generated
text, and updates that fall in silence or where the continuation is not yet
stable cost a single pre-fill. \qwen{}, in contrast, regenerates five tokens at
every update whether or not new speech has arrived, which at 160\,ms amounts
to 31.25 serial steps per second.

% ---------------------------------------------------------------------------
\subsection{Bounded Context}
\label{sec:bounded-context}

Caching removes completed audio from the recomputation but not from the
context: the cached windows and the re-pre-filled transcript still grow with
the stream. We therefore retain only the $W$ most recent completed windows in
addition to the active one, and evict each older window together with the text
committed before that window ends. Because text is always committed after the
audio it transcribes, eviction never removes the transcript of retained audio,
which would otherwise be transcribed a second time. The context is thus
limited to $8(W{+}1)$\,s of audio and its transcript. We use $W=1$, in line
with prior findings that LLM-based streaming ASR needs only limited left
context~\citep{jia2025efficientstreamingllm}. Since eviction changes the left
context of the retained window, its KV cache is rebuilt once per eviction,
i.e., once every 8\,s.

% ---------------------------------------------------------------------------
\subsection{Pre-fill Verification}
\label{sec:prefill-verification}

With $U>0$, the last $U$ generated tokens remain revisable. Instead of
regenerating them as \qwen{} does, we verify them in the pre-fill. Since the
previous suffix $r_{k-1}$ is part of the decoder input, the pre-fill already
yields the model's greedy prediction at each of its positions under the new
audio. We keep the longest prefix of $r_{k-1}$ that agrees with these
predictions, truncate the KV cache at the first disagreement, take the
replacement token from the pre-fill logits, and resume decoding. This is the
verification step of speculative
decoding~\citep{leviathan2023speculative,chen2023speculative}, with the
previous suffix as the draft. Under greedy decoding, it produces the same
output as regeneration up to numerical precision, and a suffix that is still
correct costs no decoding step. Qwen3-ASR-causal~\citep{fuxa2026qwen3asrcausal}
applies a similar check to the whole hypothesis; we restrict it to the
$U$-token suffix so that committed text is never revised. The accuracy and
latency effects of $U$ are analyzed in the next section.

\section{Evaluations}
\label{sec:evaluation}

\subsection{Experimental Setup}
\label{sec:experimental-setup}

We evaluate R2T2 on English and Chinese benchmarks and assess transfer to
ten additional languages; dataset and baseline details are provided in
Appendix~\ref{app:evaluation-data}.
The main comparison evaluates R2T2 with 160\,ms chunks under
different provisional-token budgets $U$. Committed text
remains unchanged under the policy in Section~\ref{sec:wait-terminated-decoding}.
The chunk-size sweep in Figure~\ref{fig:all-chunks-wer-latency}
uses the same R2T2 checkpoint and the same English and Chinese
cohorts. All R2T2 streaming evaluations use $W=1$, retaining one completed audio
window in addition to the active window (Section~\ref{sec:bounded-context}).
We group configurations by their output policy: stable-prefix models
preserve released text, whereas non-stable prefix models expose
revisable hypotheses.
In the 2\,s Qwen3-ASR-1.7B reference in
Table~\ref{tab:recognition-results}, all five rollback tokens
($U=5$) remain visible and can be revised as more audio arrives.
Figures~\ref{fig:bilingual-wer-latency} and~\ref{fig:all-chunks-wer-latency} also include a base run at
2\,s with $U=5$ that withholds the provisional suffix and returns only committed text, like R2T2.

\noindent\textbf{Recognition Accuracy.}
We report English WER and Chinese CER after standard text normalization.
Macro averages give equal weight to each dataset; the total macro
average combines error percentages across all 13 main test sets,
rather than pooling word and character edit counts.

\begin{figure}[H]
\centering
\begin{minipage}{0.80\linewidth}
\centering
\definecolor{latInk}{HTML}{182D43}
\definecolor{latTeal}{HTML}{007F83}
\definecolor{latTealLight}{HTML}{E1F3F1}
\definecolor{latAmber}{HTML}{A85A0A}
\definecolor{latAmberLight}{HTML}{FFF0D8}
\definecolor{latGray}{HTML}{64748B}
\resizebox{\linewidth}{!}{%
\begin{tikzpicture}[x=1cm,y=1cm,
    every node/.style={font=\sffamily\small,text=latInk}]
\path[use as bounding box] (0,-0.05) rectangle (14.5,5.2);
\foreach \xx in {4.16,6.32,8.48,10.64,12.8}
    \draw[latGray!20,dashed] (\xx,0.95) -- (\xx,4.03);
\fill[latTealLight,rounded corners=3pt] (9.65,3.92) rectangle (13.8,5.05);
\node[anchor=west,font=\sffamily\bfseries] at (0.3,4.52) {Stream};
\foreach \xx/\front/\word in {6.32/0.32/cat,8.48/0.48/dog,10.64/0.64/cat,12.8/0.80/cat} {
    \node[text=latGray,font=\sffamily\footnotesize] at (\xx,4.84) {\front};
    \node[font=\ttfamily\scriptsize] at (\xx-0.3,4.39) {a blue};
}
\foreach \xx/\word in {6.32/cat,8.48/dog} {
    \node[fill=latAmberLight,text=latAmber,rounded corners=2pt,
        font=\ttfamily\scriptsize\bfseries,inner sep=3pt]
        at (\xx+0.6,4.39) {\word};
}
\foreach \xx in {10.64,12.8}
    \node[fill=white,text=latTeal,rounded corners=2pt,
        font=\ttfamily\scriptsize\bfseries,inner sep=3pt]
        at (\xx+0.6,4.39) {cat};
\node[text=latTeal,font=\sffamily\scriptsize] at (12.8,3.84) {final};
\foreach \yy/\word/\ending/\front/\xe/\xf/\mid/\lag in {
    3.25/a/0.10/0.32/3.35/6.32/4.835/0.22,
    2.43/blue/0.24/0.32/5.24/6.32/5.78/0.08,
    1.61/cat/0.40/0.64/7.40/10.64/9.02/0.24} {
    \node[anchor=west,font=\ttfamily\small\bfseries] at (0.3,\yy) {\word};
    \draw[latGray!15] (2.0,\yy) -- (13.5,\yy);
    \draw[->,latTeal,line width=1.2pt] (\xe+0.1,\yy) -- (\xf-0.12,\yy);
    \draw[latInk,fill=white,line width=1pt] (\xe,\yy) circle (2.8pt);
    \fill[latTeal] (\xf-0.065,\yy-0.065) rectangle (\xf+0.065,\yy+0.065);
    \node[text=latTeal,font=\sffamily\small\bfseries] at (\mid,\yy+0.27) {\lag\,s};
    \node[text=latGray,font=\sffamily\scriptsize] at (\xe,\yy-0.27) {\ending};
    \node[text=latTeal,font=\sffamily\scriptsize] at (\xf,\yy-0.27) {\front};
}
\draw[->,latGray,line width=0.6pt] (2.0,0.87) -- (14.0,0.87);
\foreach \xx/\tt in {2.0/0,4.16/0.16,6.32/0.32,8.48/0.48,10.64/0.64,12.8/0.80} {
    \draw[latGray] (\xx,0.81) -- (\xx,0.93);
    \node[text=latGray,font=\sffamily\scriptsize] at (\xx,0.6) {\tt};
}
\node[anchor=east,text=latGray,font=\sffamily\scriptsize] at (14.35,0.6) {s};
\draw[latInk,fill=white,line width=1pt] (2.12,0.15) circle (2.8pt);
\node[anchor=west,font=\sffamily\footnotesize] at (2.4,0.15) {Qwen-aligned end};
\fill[latTeal] (7.75,0.085) rectangle (7.88,0.215);
\node[anchor=west,font=\sffamily\footnotesize] at (8.05,0.15) {Fixed};
\draw[->,latTeal,line width=1.1pt] (10.1,0.15) -- (10.85,0.15);
\node[anchor=west,font=\sffamily\footnotesize] at (11.05,0.15) {Lag};
\end{tikzpicture}%
}
\end{minipage}
\caption{Retrospective lag from Qwen-aligned reference word ends (circles) to the end times when the corresponding transcript prefixes become permanently stable (squares).}
\label{fig:latency-definition}
\end{figure}
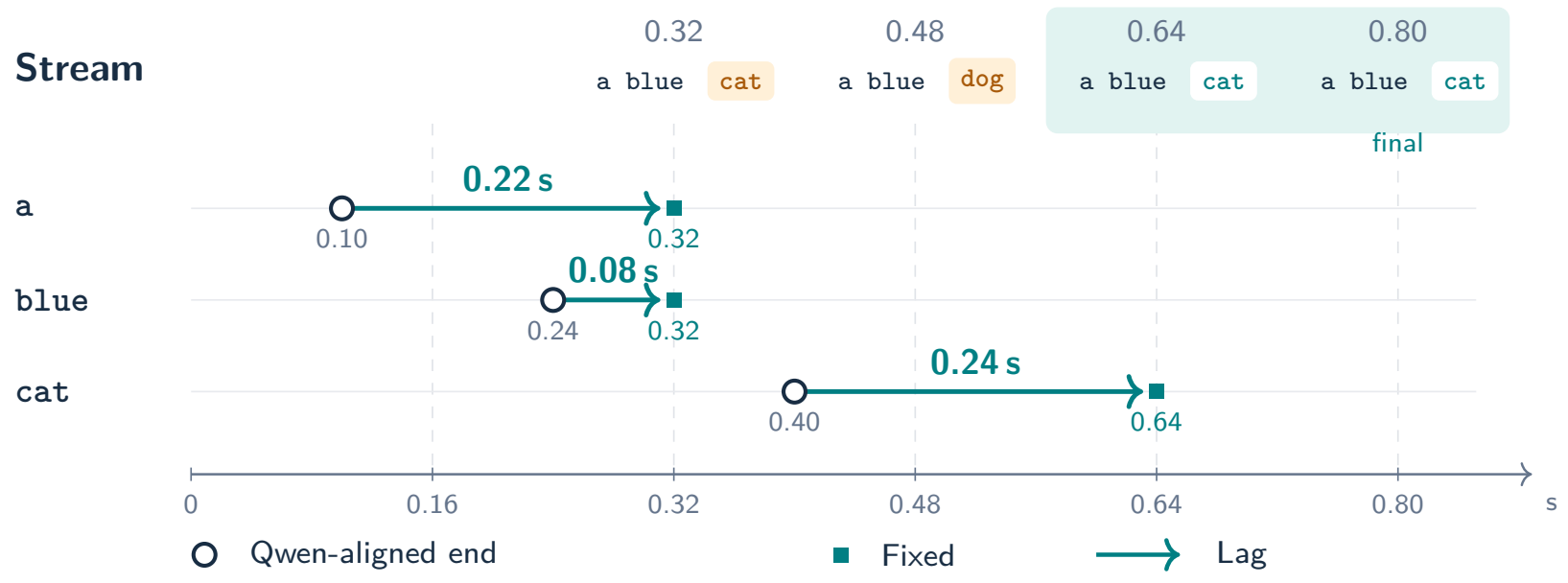

\noindent\textbf{Streaming Latency.}
For R2T2, we score normalized committed updates; the final streaming
transcript may differ from a full-context decode. Only identical units (English
words or Chinese characters) matched by Levenshtein alignment contribute to retrospective emission lag, so it
should be interpreted alongside WER/CER. For each pair, the lag is the audio
time when its output prefix first becomes permanently stable minus its
Qwen force-aligned reference end time (Figure~\ref{fig:latency-definition}).
We exclude utterances with high recognition error from lag scoring. 

The lag is measured on the audio timeline and excludes execution, queueing,
transport, and rendering. It characterizes transcript responsiveness
when inference keeps pace with incoming audio, a condition not verified
by runtime measurements here. For APIs, we estimate each update's audio time from the amount of audio
supplied; these estimates are approximate because server-side buffering is undisclosed. 

\subsection{Recognition Quality and Model Comparison}
\label{sec:results}

\begin{figure}%[H]
\centering
\begin{minipage}{0.85\linewidth}
\centering
\input{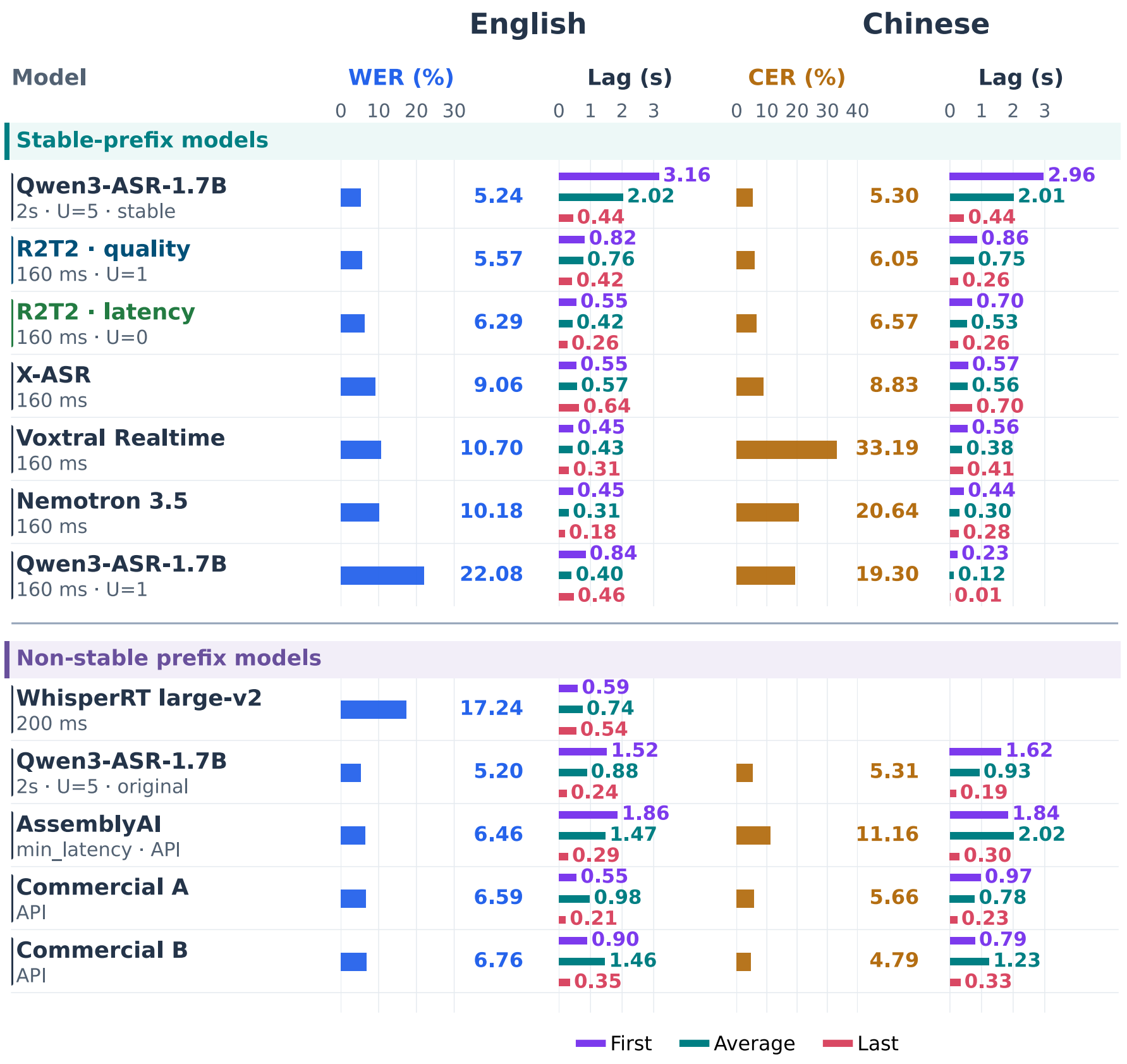}
\end{minipage}
\caption{English WER, Chinese CER, and retrospective emission lag for the main streaming configurations and 
the two 2\,s, $U=5$ base variants with different output policies. Within each dataset, Average is the mean lag; First and Last average the lags of the first and last matched units.}
\label{fig:bilingual-wer-latency}
\end{figure}

Table~\ref{tab:recognition-results} reports the main recognition
comparison, lists chunk sizes, and marks revisable intermediate
hypotheses with an asterisk (*). Both R2T2 columns use 160\,ms chunks:
$U=1$ prioritizes recognition quality, whereas $U=0$ prioritizes lower
emission latency. The base model's full-context decode, obtained by
decoding the complete utterance, provides an offline WER/CER
reference. The 2\,s, $U=5$ base configuration provides a strong
accuracy reference, but with higher emission lag (Figure~\ref{fig:bilingual-wer-latency}).

\begin{table}[H]
\centering
\footnotesize
\setlength{\tabcolsep}{3pt}
\renewcommand{\arraystretch}{1.16}
\resizebox{\linewidth}{!}{%
\begin{tabular}{@{}l*{12}{c}@{}}
\toprule
Dataset & \multicolumn{9}{c}{Open-source} & \multicolumn{3}{c}{Proprietary} \\
\cmidrule(lr){2-10}\cmidrule(lr){11-13}
& Base & Base* & Base & \multicolumn{2}{c}{\textbf{R2T2}} & X-ASR & WhisperRT* & \shortstack{Nemotron\\3.5} & \shortstack{Voxtral\\Realtime} & AssemblyAI* & Comm.~A* & Comm.~B* \\
& Offline & 2\,s & 160\,ms & 160\,ms & 160\,ms & 160\,ms & 200\,ms & 160\,ms & 160\,ms & \texttt{min\_latency} & & \\
& & $U=5$ & $U=1$ & $U=1$ & $U=0$ & & & & & & & \\
\midrule
\multicolumn{13}{@{}l}{\textbf{English (WER, \%)}} \\
\addlinespace[2pt]
AMI & 9.18 & 9.25 & 24.79 & \underline{10.07} & 10.40 & 14.41 & 24.19 & 18.11 & 17.58 & 12.00 & 13.27 & \textbf{\underline{8.44}} \\
Giga-clean & 8.60 & 8.61 & 24.37 & \textbf{\underline{8.81}} & 9.04 & 10.26 & 13.81 & 12.67 & 12.16 & 9.21 & \underline{8.84} & 9.46 \\
LS-clean & 1.62 & 1.67 & 18.37 & \underline{1.91} & 2.12 & 3.86 & 4.70 & 3.71 & 3.05 & 1.89 & 1.73 & \textbf{\underline{1.25}} \\
LS-other & 3.39 & 3.54 & 25.74 & \underline{4.12} & 4.55 & 9.64 & 9.86 & 8.27 & 8.70 & 3.37 & 3.57 & \textbf{\underline{2.48}} \\
VoxPopuli & 3.12 & 3.02 & 20.71 & \textbf{\underline{2.29}} & 3.54 & 5.68 & 8.28 & 5.69 & 8.00 & 4.75 & 3.17 & \underline{3.14} \\
Earnings22 & 6.26 & 6.68 & 29.72 & \underline{7.61} & 9.06 & 15.95 & 35.08 & 17.22 & 12.12 & \textbf{\underline{7.47}} & 10.32 & 8.96 \\
TED-LIUM & 2.26 & 2.33 & 19.18 & \textbf{\underline{2.46}} & 2.77 & 3.75 & 6.67 & 5.11 & 5.89 & 3.23 & \underline{3.08} & 3.30 \\
EN-RealSI & 6.84 & 6.54 & 13.76 & \textbf{\underline{7.33}} & 8.84 & 8.97 & 35.36 & 10.69 & 18.08 & 9.73 & \underline{8.73} & 17.05 \\
\addlinespace[4pt]
\midrule
\multicolumn{13}{@{}l}{\textbf{Chinese (CER, \%)}} \\
\addlinespace[2pt]
Wenet-net & 4.87 & 4.94 & 19.79 & \underline{5.75} & 6.17 & 8.81 & N/A & 24.70 & 27.36 & 12.91 & 5.13 & \textbf{\underline{4.79}} \\
Wenet-meeting & 5.85 & 5.97 & 20.38 & \underline{6.74} & 7.34 & 11.33 & N/A & 20.18 & 66.50 & 11.84 & 7.07 & \textbf{\underline{3.75}} \\
SPEECHIO-06 & 5.90 & 6.10 & 24.47 & \underline{7.19} & 7.43 & 7.86 & N/A & 22.52 & 35.33 & 15.08 & 5.67 & \textbf{\underline{5.34}} \\
SPEECHIO-07 & 6.15 & 6.19 & 21.16 & \underline{7.29} & 7.94 & 11.22 & N/A & 24.28 & 26.84 & 10.84 & \textbf{\underline{6.45}} & 6.46 \\
CN-RealSI & 2.91 & 3.34 & 10.72 & \textbf{\underline{3.27}} & 3.98 & 4.92 & N/A & 11.52 & 9.92 & 5.15 & 3.99 & \underline{3.64} \\
\midrule
Total macro average & 5.15 & 5.25 & 21.01 & \textbf{\underline{5.76}} & 6.40 & 8.97 & N/A & 14.21 & 19.35 & 8.27 & 6.23 & \underline{6.01} \\
\bottomrule
\end{tabular}%
}
\caption{English WER and Chinese CER (\%). Base denotes Qwen3-ASR-1.7B; Offline uses full-context decoding. The total macro average combines English and Chinese error rates with equal weight for each of the datasets. Within each row, bold marks the lowest error; underlining marks the lowest error within each open-source or proprietary group. Base Offline and Base 2\,s ($U=5$) are reference-only and excluded from both highlighting comparisons. * denotes revisable partial hypotheses; N/A denotes unavailable results.}
\label{tab:recognition-results}
\end{table}

\paragraph{Adaptation and Recognition Quality.}
The matched comparison in Table~\ref{tab:recognition-results}
shows that R2T2 develops strong recognition with small audio
chunks. At 160\,ms and $U=1$, it improves on the base model on
every listed English and Chinese test set, reducing the total macro
recognition error by nearly three quarters. For unadapted Qwen3-ASR-1.7B at 160\,ms, initial $U=0$
runs showed severe hallucination and unreliable termination,
preventing usable recognition-error evaluation. The main table
therefore reports only $U=1$ for this baseline at 160\,ms.
R2T2 also operates with $U=0$ at this chunk duration. These findings suggest that R2T2 has learned to transcribe
partial audio and wait for additional context when needed,
consistent with the LSP supervision described in Section~\ref{sec:lsp-supervision}.

R2T2's streaming recognition closes most of the gap to the base
model's full-context performance, leaving a modest difference of
about 0.6 percentage points.
Some gap is expected, as resolving certain ambiguities requires
future context unavailable when streaming output is committed.
On VoxPopuli, it even outperforms the full-context reference.
R2T2 therefore delivers accurate recognition from small audio
chunks with stable-prefix output: once text is committed, it remains unchanged.
Figure~\ref{fig:bilingual-wer-latency} further relates
this recognition quality to low emission lag.

\paragraph{Overall Comparison and Practical Applicability.}
Among the configurations ranked in
Table~\ref{tab:recognition-results}, R2T2 with $U=1$ has the
lowest total average error of 5.76, surpassing even all three evaluated
proprietary APIs. It also leads the eligible open-source
configurations on all test sets at the
evaluated 160\,ms settings. These rankings exclude the full-context
and 2\,s base references.
R2T2's overall lead is reinforced by its streaming responsiveness:
it achieves lower mean emission lag than all three proprietary
APIs in both languages, alongside lower English macro error
(Figure~\ref{fig:bilingual-wer-latency}).
The Chinese accuracy advantage of Commercial A and Commercial B
is accompanied by later transcript stabilization, highlighting
the importance of assessing accuracy and responsiveness together.

Chinese-language support alone does not ensure competitive
recognition: Nemotron 3.5 and Voxtral Realtime support Chinese but perform
substantially worse on this cohort, with Voxtral Realtime's Wenet-meeting
result also showing sensitivity to the speech domain. WhisperRT
does not support Chinese, so its Chinese entries are unavailable.
R2T2's lead among the ranked configurations on both RealSI
language cohorts is particularly relevant to practical streaming
use. RealSI contains long-form
speech from real-world simultaneous interpretation, bringing
the evaluation closer to sustained speech use cases than isolated
utterance tests. Strong recognition on this benchmark, combined
with committed output, highlights R2T2's applicability to streaming
tasks that consume text incrementally without repeatedly replacing
earlier content.

\medskip
\noindent\textbf{Cross-Language Transfer.}
Although R2T2 is post-trained only on Chinese and English, its streaming
gains extend to 10 additional languages: at 160\,ms and $U=1$, recognition
error is lower than the base model's at the same setting on all 18 test
sets (Table~\ref{tab:language-transfer} in Appendix~\ref{app:language-transfer}). 

\subsection{Emission Lag and Inference Trade-offs}
\label{sec:latency-results}

\noindent\textbf{Stability and Recognition--Latency Trade-off.}
Figure~\ref{fig:bilingual-wer-latency} illustrates a recognition--lag
trade-off: among the short-chunk baselines, stable-prefix models tend to
offer lower emission lag at higher recognition error. Non-stable prefix
models, including the evaluated proprietary APIs, can use later context
to correct earlier hypotheses, often achieving better final WER/CER but
stabilizing later, as seen in the Chinese results of Commercial A and
Commercial B. Changes to displayed text cause transcript flicker and may
require downstream reprocessing
\citep{mcgraw2012wordstability,bruguier2023flickering}. The two
Qwen3-ASR-1.7B variants at 2\,s and $U=5$ illustrate the cost of requiring
stable output: both achieve similarly strong final recognition, but
withholding provisional text produces substantially higher mean and
initial lag than displaying revisable hypotheses.

R2T2 narrows this gap by combining strong recognition quality with emission
lag comparable to the short-chunk stable-prefix baseline. Its explicit
commitment boundary supports immutable incremental output, whereas
applications receiving revisable text need an additional stability or
finalization policy \citep{machacek2023whisperstreaming}. Retrospective lag
reflects revisions that change earlier text through later prefix fixation,
while unchanged provisional text receives credit from its first appearance.
Permitting revisions therefore does not invariably increase this metric,
and flicker frequency itself is not measured here. R2T2 uses \waittoken{}
to end decoding for the current audio update; detecting the end of speech
and flushing the final output remain separate runtime operations.

\FloatBarrier
\begin{figure}[H]
\centering
\begin{minipage}{0.90\linewidth}
\centering
\input{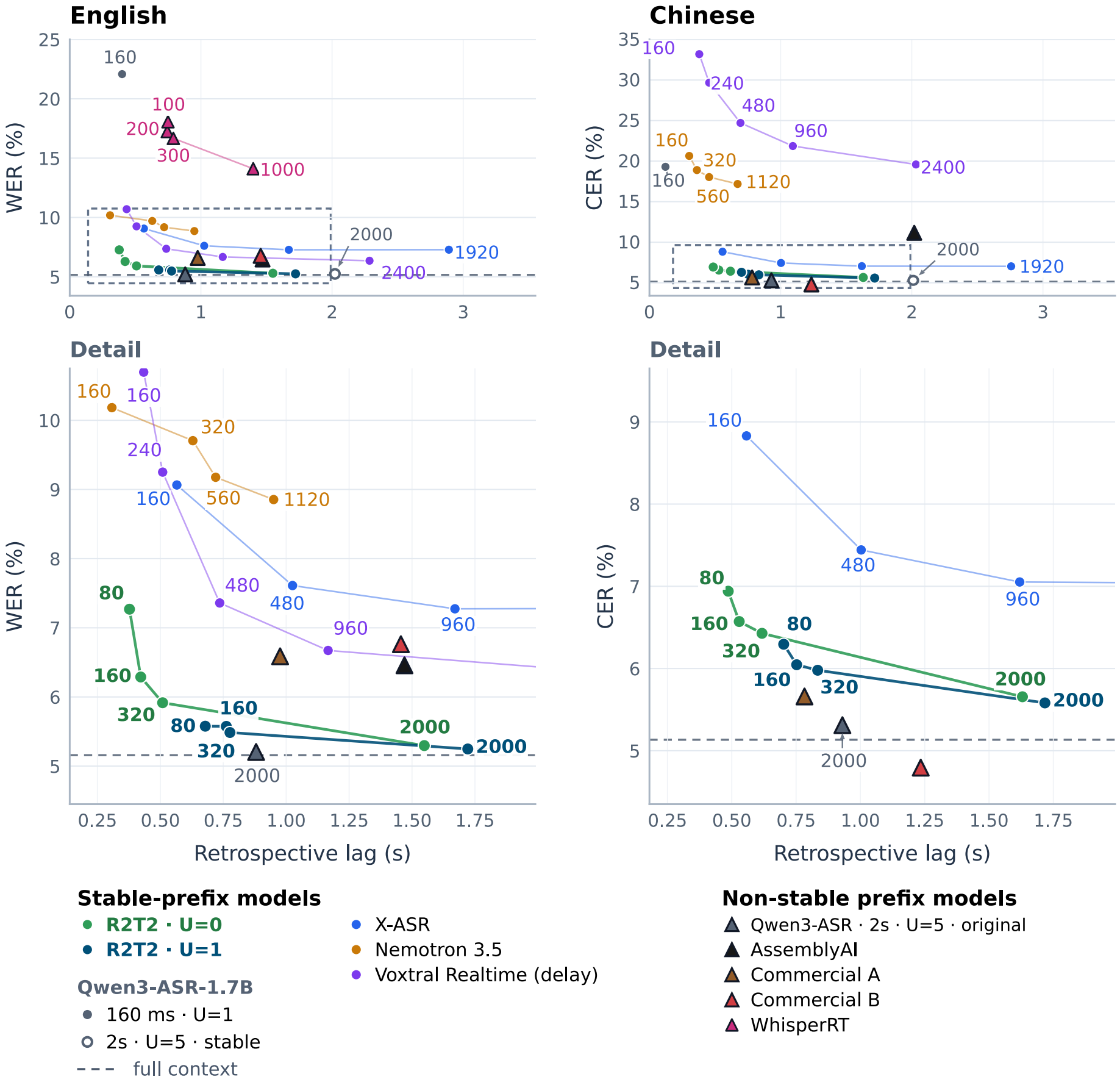}
\end{minipage}
\caption{Recognition error and retrospective emission lag across
chunk sizes, decoding delays, and API settings. Triangles denote non-stable prefix
models.}
\label{fig:all-chunks-wer-latency}
\end{figure}

\medskip
\noindent\textbf{Chunk-Size and Provisional-Suffix Sensitivity.}
Figure~\ref{fig:all-chunks-wer-latency} compares recognition error and mean
retrospective emission lag across chunk sizes and provisional-token budgets.
Curves connect measured settings; dashed lines indicate the base model's
full-context recognition error. R2T2's budget $U$ controls the recognition--lag
trade-off at inference without changing the checkpoint or requiring further
training. Both modes release only committed text, keeping the provisional
suffix internal. At 160\,ms, withholding one token reduces total macro
recognition error by about one tenth while adding a few tenths of a second
to mean lag. Even with $U=0$, R2T2 maintains lower macro error in both languages
than the other open-source baselines evaluated at 160\,ms, although some
achieve lower lag with substantially higher error. Withholding a token also
reduces sensitivity to chunk duration: doubling it from 160\,ms to 320\,ms
brings little additional recognition benefit with $U=1$, whereas $U=0$ is
more sensitive. This is consistent with the provisional suffix allowing
additional context to arrive before commitment. At 2\,s, the larger chunks
already provide more context, so withholding one token brings only small
changes in recognition quality and lag. Compared with 160\,ms, however,
the modest recognition improvement comes with substantially higher emission lag.

Chunk duration alone therefore does not determine responsiveness. Smaller
chunks do not always produce lower lag because the timing and distribution
of speech, including word boundaries and pauses, affect when enough context
becomes available for commitment. Waiting for that context can delay output
even when the real-time factor (RTF) is below one. Nevertheless, 80\,ms chunks
with $U=0$ provide a latency-priority setting, keeping mean retrospective lag
below half a second in both languages at some cost to recognition quality.
These values describe emission timing on the audio timeline rather than
measured serving latency (Section~\ref{sec:experimental-setup}). The baseline
configurations illustrate a similar trade-off: providing more audio context
can improve recognition while delaying stable output. Non-stable prefix
models, represented by triangles in Figure~\ref{fig:all-chunks-wer-latency},
can also achieve strong final recognition, but revisions to earlier text
delay prefix fixation, making their retrospective lag less competitive
than R2T2's short-chunk modes.

\medskip
\noindent\textbf{Tokenization Considerations.}
A fixed provisional-token budget does not correspond to a fixed delay in
milliseconds: the duration of the speech represented by those tokens varies
with speaking rate, language, and tokenization. With Qwen3-ASR-1.7B's byte-pair encoding
(BPE) tokenizer, the text represented by one token can be a prefix of the
text represented by another. Consequently, agreement between text prefixes
does not necessarily imply agreement between decoder token sequences.
This distinction requires care when constructing streaming supervision
and implementing inference.

\section{Conclusion}

\sysname{} shows that a strong full-context ASR model can be adapted to accurate streaming
recognition while retaining its pre-trained audio encoder and projector and preserving released text. LSP
supervision and explicit waiting targets train the model to recognize partial audio and
defer uncertain continuations. Across the evaluated English and Chinese benchmarks, \method{} closes most of the
recognition gap to the base model's full-context inference while providing low retrospective
emission lag and leading the ranked configurations in combined macro error. Its transfer
gains extend beyond the two post-training languages. Configurable chunk duration and
provisional-token budget expose different recognition--lag trade-offs without retraining, while
cache reuse, bounded context, and suffix verification limit repeated inference work. These
findings support \method{} as a practical foundation for applications that consume stable text
incrementally, with deployment performance and broader generalization remaining priorities
for further evaluation.

\newpage
\bibliographystyle{plainnat}
\bibliography{references}

\newpage
\appendix

\section{The Longest Stable Prefix as a Learning Target}\label{sec:lsp-learning}

To the best of our knowledge, there are already two definitions in the literature under the name ``stable prefix''.
The two, together with ours, are not the same object; there is no total order
among them (the final-anchored formulation contains the other two, while the
other two straddle each other), and they coincide only on trajectories that are
already append-only. The differences between them are exactly the differences
that matter when the
stable prefix is used as supervision rather than as a diagnostic. We state all
three, establish their exact relations
(Section \ref{sec:lsp-relations}), and then identify what our formulation is
maximal for (Section \ref{sec:lsp-why}).

We use the notation introduced in Section~\ref{sec:lsp-notation}. We first highlight the following:

\begin{enumerate}
\item \emph{Two different ``futures''.} The material available for judging a
prefix at time $t$ is either the \emph{trajectory} $\{y_{t'}\}_{t' \ge t}$ that
the same decoding run goes on to produce, or the model's own
\emph{full-context output} $\bar y$. The first is a fact about the path this run
took, the second a fact about what the same model would produce with the whole
utterance in view.
Definition~\ref{def:allfuture} anchors to the first, Definitions~\ref{def:finalanchored}
and~\ref{def:lsp} to the second.
\item \emph{All three are hindsight quantities.} Each needs material that does
not exist at $t$, so none of them is observable online. This is not a defect of
one formulation: it is what makes all three usable as \emph{offline labels},
and it is why the applied literature estimates rather than measures stability
online \citep{mcgraw2012wordstability}. The difference between the three is
therefore \emph{what} a label certifies, not \emph{when} it can be computed.
\item \emph{Accuracy is not part of the definitions.} All three are anchored, directly or
indirectly, to the model's own full-context output rather than a ground truth transcription.
They constrain \emph{when} the model may commit, not
\emph{what} it should say. This is deliberate. Accuracy already has calibrated instruments
--- WER, CER, and their domain-specific variants --- but streaming stability does not. 
Keeping the target self-referential makes it a statement about
commit timing alone, so that a streaming model can be compared with its own
non-streaming self on a single scale: the definition says nothing about the gap
between that self and the truth, and WER measures exactly that gap.

\end{enumerate}

\subsection{Three Formulations}\label{sec:lsp-defs}

We present them as three different \emph{maximizations} over the same material,
not as a ladder of increasing strength; Section \ref{sec:lsp-relations} shows that no
such ladder exists.

\begin{definition}[All-future criterion; \citealp{mcgraw2012wordstability}]
\label{def:allfuture}
Let $y_1, \dots, y_T$ be the incremental results actually produced as the
utterance is decoded, and let $\mathcal{F}_t = \{y_{t'} : t' \ge t\}$ be the
later results of that same run, $y_t$ included. A prefix of $y_t$ is
\emph{stable} at $t$ if every member of $\mathcal{F}_t$ shares it. Its length is
\begin{equation}
  k^{\star}(t) \;=\; \max\Bigl\{\, k \le n_t \;:\; \forall\, y' \in
  \mathcal{F}_t,\; y'_{1..k} = y_{t,1..k} \,\Bigr\},
  \label{eq:kstar}
\end{equation}
and the \emph{stable prefix} is $y_{t,1..k^{\star}(t)}$. The set is non-empty
since $k = 0$ always qualifies, and the maximum is attained because agreement on
$1..k$ implies agreement on $1..k-1$.
\end{definition}

Three points about Definition~\ref{def:allfuture} are easy to get wrong, and
each matters later.

\begin{enumerate}
\item $\mathcal{F}_t$ is a \emph{realized} sequence, not a set of alternatives.
It contains the outputs the run does produce at later frames, and the criterion
is agreement among them: a prefix is stable ``only if all future incremental
results have the same prefix'' \citep{mcgraw2012wordstability}. It is not a set
of hypotheses reachable from $x_{\le t}$, and $k^{\star}(t)$ is therefore not a
function of $x_{\le t}$ alone.
\item It is an \emph{oracle}, computable only after decoding. The paper is
explicit that the ideal prefixes are not available online and must be estimated
by a predictor \citep{mcgraw2012wordstability}.
\item Its agreement with $\bar y$ is \emph{implied} rather than checked
separately. Since $x_{\le T} = x$, we have $y_T = \mathcal{M}(x_{\le T}) =
\mathcal{M}(x) = \bar y$, so the endpoint of the run \emph{is} the full-context
output and any prefix stable in the all-future sense is automatically a prefix
of $\bar y$ as well: the criterion is not $\bar y$-blind. What it \emph{adds} is
agreement with the \emph{intermediate} outputs: a prefix shared by $y_t$ and
$\bar y$ still fails if some $y_{t'}$, $t < t' < T$, departs from it even
though a later output returns. In that sense it certifies durability along the
run's own path, which no endpoint-only criterion tests.
\end{enumerate}

\begin{definition}[Final-anchored longest common prefix; \citealp{hoang2026dynamic}]
\label{def:finalanchored}
The stable prefix at step $t$ is
\begin{equation}
  m_t^{\mathrm{FA}} \;=\; \max_{j \le t} \ell_j, \qquad
  L_t^{\mathrm{FA}} \;=\; \bar y_{1..m_t^{\mathrm{FA}}},
  \label{eq:FA}
\end{equation}
with $\ell_j$ as in \eqref{eq:ell}.
\end{definition}

The running maximum makes the target monotone under the prefix order by construction: content that has
agreed with the full transcript and been committed cannot be retracted by a
noisier later partial result \citep{hoang2026dynamic}. This is a partial-credit
criterion: an output that agrees with $\bar y$ on a head and diverges on the tail
contributes its agreeing head. It is consistent with $\bar y$ by construction,
and, being
a maximum over $\ell_j$, it can name a prefix that no single output of the model
ever realized.

\begin{definition}[Model-admissible final-anchored prefix; \emph{ours}]
This formulation was introduced in Definition~\ref{def:lsp} (Section~\ref{sec:lsp_learning_main}). Write $L_t = \LSP(\mathcal{M},\, x,\, t)$ for conciseness.
\end{definition}

$L_t$ is \emph{the longest prefix of its own full-context output that the model has already produced
from audio available at time $t$}. Definition~\ref{def:lsp} therefore keeps two
commitments at once. It is anchored to the \emph{final} full-context output, which
excludes targets that this output contradicts; and it is an object the model has
\emph{actually generated}, which excludes targets that exceed what the model can
decode from $x_{\le t}$.

We call a label \emph{model-admissible} at $t$ when it \emph{equals} an output
the model produced from audio available at $t$, i.e.\ $y_j$ for some $j \le t$;
a truncation of one output, or a string assembled from several, is not
model-admissible. This is the strict reading used in the rest of this appendix.
Under the weak reading, in which the label need only be a \emph{prefix} of an
earlier output, all three formulations would qualify and the corresponding
column of Table~\ref{tab:lsp-compare} would not discriminate.

\begin{example}[The three differ off the ideal trajectory]
\label{ex:threeway}
Let $\bar y = $ ``a b c d e'' and let the run produce $y_1 = $ ``a b c x'',
$y_2 = $ ``a b'', $y_3 = $ ``a b c d e''. At $t = 2$ the futures are $y_2$ and
$y_3$, which share ``a b'', so Definition~\ref{def:allfuture} gives
$k^{\star}(2) = 2$, truncated by the short current output.
Definition~\ref{def:finalanchored} gives $m_2^{\mathrm{FA}} = \max(3, 2) = 3$,
through $y_1$: an agreeing head on a diverging tail, giving a string that no
output of the run equals. Definition~\ref{def:lsp} gives $m_2 = 2$, since $y_2
\prefeq \bar y$ while $y_1$ does not. At $t = 3$, where $y_3 = \bar y$ and
the outputs already form a chain, all three give five.
\end{example}

The three coincide on the trajectory one would like a recognizer to have, which
is why a single name has survived; they diverge exactly where a training signal
is needed. We make this precise as Proposition~\ref{prop:coincide}.

\subsection{Relations among the Three}\label{sec:lsp-relations}

\begin{proposition}[Ours is contained in the final-anchored prefix]
\label{prop:contain}
$m_t \le m_t^{\mathrm{FA}}$ for all $t$, i.e.\ $L_t \prefeq L_t^{\mathrm{FA}}$.
\end{proposition}
\begin{proof}
Every admissible term in \eqref{eq:mt} is $\ell_j$ for some $j \le t$, since
$y_j \prefeq \bar y$ gives $\ell_j = n_j$; a maximum over a subset cannot
exceed the maximum over $\{\ell_j\}_{j \le t}$.
\end{proof}

\begin{proposition}[The all-future prefix is contained in the final-anchored
prefix]\label{prop:allfuture-contain}
For every $t$, $k^{\star}(t) \le \ell_t \le m_t^{\mathrm{FA}}$.
\end{proposition}
\begin{proof}
Since $x_{\le T} = x$, $y_T = \mathcal{M}(x) = \bar y$, so $y_T \in
\mathcal{F}_t$ without further hypothesis. The stable prefix is shared with
$y_T$, hence $y_{t,1..k^{\star}(t)} = y_{T,1..k^{\star}(t)} =
\bar y_{1..k^{\star}(t)}$ and $k^{\star}(t) \le \ell_t$; and $\ell_t \le
m_t^{\mathrm{FA}}$ by \eqref{eq:FA}.
\end{proof}

The containment is strict in general: the all-future label can be shorter than
$\ell_t$, because agreement with an intermediate output is not implied by
agreement with the endpoint (Proposition~\ref{prop:incomparable}). What it buys
in exchange is that its tokens have survived every output the run produced, not
merely the last one.

\begin{proposition}[The all-future prefix and ours are incomparable]
\label{prop:incomparable}
Neither $k^{\star}(t) \le m_t$ nor $m_t \le k^{\star}(t)$ holds in general.
\end{proposition}
\begin{proof}
Two witnesses.

\emph{All-future exceeds ours.} Let $\bar y = $ ``a b c d'' and let the run
produce $y_1 = $ ``a b c x'', $y_2 = $ ``a b c y'', $y_3 = $ ``a b c d''. At
$t = 2$ the futures $y_2$ and $y_3$ share ``a b c'', so $k^{\star}(2) = 3$, while
$m_2 = 0$ in \eqref{eq:mt}, since neither $y_1$ nor $y_2$ is a prefix of
$\bar y$. Note also that ``a b c'' equals no output of the run: a prefix stable
in the all-future sense is a truncation of $y_t$, not necessarily a string the
model ever produced.

\emph{Ours exceeds all-future.} Let $\bar y = $ ``a b c d'' and let the run
produce $y_1 = $ ``a b c d'', $y_2 = $ ``a b'', $y_3 = $ ``a b c d''. Then
$k^{\star}(2) = 2$, the length of the shortest output the futures share,
whereas $m_2 = 4$: the maximum in \eqref{eq:mt} ranges over \emph{past} outputs
and can therefore exceed the length of the current one, while $k^{\star}(t)$ is
capped by $n_t$ in \eqref{eq:kstar}.
\end{proof}

\begin{proposition}[Coincidence on an append-only trajectory]
\label{prop:coincide}
Suppose $y_t \prefeq y_{t'}$ for all $t \le t'$. Then for
every $t$,
\[
  k^{\star}(t) \;=\; n_t \;=\; \ell_t, \qquad m_t \;=\; m_t^{\mathrm{FA}} \;=\; n_t,
\]
so all three formulations name the whole current output.
\end{proposition}
\begin{proof}
Every member of $\mathcal{F}_t$ extends $y_t$, so $k^{\star}(t) \ge n_t$, and
$k^{\star}(t) \le n_t$ by \eqref{eq:kstar}. Since $y_t \prefeq y_T = \bar y$,
$\ell_t = n_t$ and each term of \eqref{eq:FA} is at most $\ell_t = n_t$ while the
term $j = t$ attains it. For \eqref{eq:mt}, every term is at most $n_t$ and the
term $j = t$ equals $n_t$.
\end{proof}

\begin{proposition}[The converse fails for the final-anchored formulation]
\label{prop:converse}
$m_t = n_t$ for all $t$ does \emph{not} imply that the outputs form a prefix
chain; $k^{\star}(t) = n_t$ for all $t$ does.
\end{proposition}
\begin{proof}
For the first claim, take $y_1 = $ ``a b c'', $y_2 = $ ``a b x'', $y_3 = $ ``a b
c d'' with $\bar y = $ ``a b c d'': $m_t = n_t = 3$ at $t = 1, 2$ (through
$y_1$) and $m_3 = n_3 = 4$, yet $y_1$ is not a prefix of $y_2$. For the second, $k^{\star}(t) = n_t$ says every $y_{t'}$ with $t' \ge t$
agrees with $y_t$ on all of its $n_t$ positions, i.e.\ $y_t \prefeq y_{t'}$.
\end{proof}

Propositions~\ref{prop:contain}--\ref{prop:converse} settle the relations. There
is no total order: Definition~\ref{def:finalanchored} contains the other two
(Propositions~\ref{prop:contain} and~\ref{prop:allfuture-contain}), while
Definitions~\ref{def:allfuture} and~\ref{def:lsp} straddle each other in both
directions (Proposition~\ref{prop:incomparable}). They coincide precisely on
trajectories that are already append-only, so ``the stable prefix'' is
unambiguous only where nothing has to be learned.

\subsection{Properties of Definition~\ref{def:lsp}}\label{sec:lsp-props}

\begin{proposition}[Monotonicity]\label{prop:mono}
For all $t \le t'$, $L_t \prefeq L_{t'}$; no committed stable prefix is ever
retracted.
\end{proposition}
\begin{proof}
Each term of the maximum in \eqref{eq:mt} is non-negative and the index set
$\{j : j \le t\}$ grows with $t$, so $m_t$ is nondecreasing. As $L_t$ and $L_{t'}$
are both prefixes of $\bar y$, $m_t \le m_{t'}$ implies $L_t \prefeq L_{t'}$.
\end{proof}

Monotonicity here is a property of the \emph{label}, and it is worth separating
from the monotonicity of the other two: Definition~\ref{def:allfuture}'s oracle
is monotone under the prefix order because the family $\mathcal{F}_t$ shrinks with $t$ (agreement with a
smaller set is easier), and Definition~\ref{def:finalanchored}'s is monotone under the prefix order by
the same running maximum as ours. Monotonicity therefore does \emph{not}
distinguish the three, and we do not use it as a point of comparison.

\begin{proposition}[Model admissibility]\label{prop:admissible}
If $m_t > 0$, then $L_t = y_j$ for some $j \le t$: the label is a transcript the
model itself produced from audio available at $t$.
\end{proposition}
\begin{proof}
If $m_t > 0$, the maximum in \eqref{eq:mt} is attained at some $j \le t$ with
$y_j \prefeq \bar y$ and $n_j = m_t$, whence $y_j = \bar y_{1..m_t} = L_t$.
\end{proof}

At $t = T$ the term $j = T$ contributes $n_T = |\bar y|$, so $m_T = |\bar y|$
and $L_T = \bar y$: the label chain ends at the full-context output, as stated
in the main text.

\begin{theorem}[Maximality]\label{thm:maximal}
Let $p$ be a candidate target at time $t$ that is (i) \emph{self-consistent},
$p \prefeq \bar y$, and (ii) \emph{model-admissible}, $p = y_j$ for some
$j \le t$. Then $|p| \le m_t$, with equality for $p = L_t$. Hence $L_t$ is the
longest target satisfying both conditions, and it is the unique string of that
length.
\end{theorem}
\begin{proof}
By (ii) $p = y_j$, and by (i) $y_j \prefeq \bar y$, so the indicator in
\eqref{eq:mt} is one for this $j$ and its term equals $n_j = |p|$; therefore
$m_t \ge |p|$. $L_t$ satisfies (i) as a prefix of $\bar y$ and (ii) by
Proposition~\ref{prop:admissible}. Uniqueness follows because $L_t$ is the only
prefix of $\bar y$ of length $m_t$.
\end{proof}

Theorem~\ref{thm:maximal} is the precise formal content of our preference for
Definition~\ref{def:lsp}. It is a claim about \emph{maximality among labels that
equal a whole earlier output}, not about optimality (partial credit can be longer;
see Table~\ref{tab:lsp-compare}) and not about consistency: consistency alone is shared with
Definition~\ref{def:finalanchored}, and being an output of the model is shared,
in a weaker sense, with Definition~\ref{def:allfuture}.

\subsection{Why This Formulation}\label{sec:lsp-why}

\paragraph{Two dichotomies, and one cell that satisfies both.}
The axes of Table~\ref{tab:lsp-compare} are independent, and each has a distinct
failure mode attached to it.

\emph{Anchor} (the run's trajectory vs.\ its endpoint). Anchoring to the
trajectory, as Definition~\ref{def:allfuture} does, certifies that the tokens
survived every output the run produced. That is the strictest requirement on
settledness among the three, and it costs label length whenever an intermediate
output departs from a prefix and a later one returns to it. Anchoring to $\bar y$
certifies only that the tokens survive the \emph{last} output, which is the
endpoint of that same trajectory. Since $y_T = \bar y$, the two tests are not
independent: they agree at $T$ and differ only in what they require of the
intermediate steps. Anchoring to $\bar y$ makes every label token one that
reappears in the model's own full-context commitment. Whether that commitment is
right is a separate question, which this definition deliberately
leaves to WER rather than folding into the target.

\emph{Model admissibility} (the label equals an entire earlier output vs.\ it is
a computation over outputs). The final-anchored label is a maximum of partial
agreements $\ell_j$. Since $\ell_j \le n_j$, it never exceeds the length of a
single output, but it need not equal any output: it credits the agreeing head of an
output whose tail the model itself revised, and how much it credits is decided by
comparing that output with the \emph{complete} $\bar y$, hence by evidence
carried by $x_{>t}$. Supervising $x_{\le t} \mapsto L_t^{\mathrm{FA}}$ therefore
asks the model to commit content whose evidence has not arrived, and to treat a
state in which it had not settled as if it had --- the failure mode we call label
leakage, the supervision-side counterpart of the right-context lag discussed in
Section~\ref{sec:challenges}.

Definition~\ref{def:lsp} is the intersection: anchored to the model's own
full-context output because it is a prefix of $\bar y$, and realizable because it
is one of the model's own streaming outputs.
Theorem~\ref{thm:maximal} shows that the intersection is not merely inhabited but
has a longest element, so the construction costs nothing in label length among
targets with both properties.

\paragraph{Self-distillation without a teacher.} Because the label $L_t$ is the
output generated by the same backbone from the same audio, conversion requires no external
aligner, no stronger teacher model and no distributional gap between the label
and what the model can reach. The labeling procedure is a filter over the
model's own outputs, which is also why it transfers to any checkpoint without
retraining the labeller.

\paragraph{Relation to a model's stable-prefix property.}
Definition~\ref{def:spmodel} and the all-future criterion
(Definition~\ref{def:allfuture}) describe the same property from two sides: by
Proposition~\ref{prop:converse}, a system is a stable-prefix model exactly when
its all-future stable prefix covers its whole emission at every step. The longest
stable prefix is therefore uninformative on a stable-prefix model's own output,
where it is simply the output, and informative only on a system that revises.
The full-context model $\mathcal{M}$ run on truncated audio is such a system:
$y_t = \mathcal{M}(x_{\le t})$ ends in a tail guessed from incomplete audio,
which later outputs may change. The longest stable prefix is thus the part of a
non-stable model's output that can be kept, and it is the label we extract from
$\mathcal{M}$ in order to train a stable-prefix model.

%\paragraph{The price.} All-or-nothing credit is strict, and three
%costs follow. (i) A single tail token that diverges from $\bar y$ zeroes the
%credit, so labels can be shorter than what is genuinely durable
%(Proposition~\ref{prop:incomparable}, first witness); since
%Definition~\ref{def:allfuture} can be longer, ``strictest of the three'' would be
%false, and the accurate statement is that ours is the smaller of the two
%\emph{full-context-anchored} formulations. (ii) The bias is towards later emission,
%hence a latency cost, which \S\ref{sec:lsp-predictor} treats as a separate axis.
%(iii) All-or-nothing credit is sensitive to how $\bar y$ happens to be written:
%one orthographic or disfluency mismatch in the middle of an utterance destroys
%every subsequent label for that utterance, where
%Definition~\ref{def:finalanchored} degrades gracefully. The obvious mitigations
%--- labeling against a normalised $\bar y$, or falling back to partial credit
%$\ell_j$ when $y_j$ is close to $\bar y$ --- weaken
%Proposition~\ref{prop:admissible} and could be future 
%directions to explore. In our setting the rates of both failure modes, (a) the fraction of steps with
%$m_t < m_t^{\mathrm{FA}}$ and (b) the mean gap $m_t^{\mathrm{FA}} - m_t$, 
%are small and Section \ref{sec:streamability} and our final evaluations establish that
%the negative effect on downstream quality is not severe. We keep the strict condition rather than relax it because it
%is what makes Theorem~\ref{thm:maximal} and
%Proposition~\ref{prop:admissible} true.

\subsection{LSP learning as the paradigm for stable-prefix streaming recognition}\label{sec:lsp-predictor}

We now connect the target to the behaviour it is meant to induce.
Recall $y_t=\mathcal{M}(x_{\le t})$, the model's output on the clip heard so far.

\paragraph{The ideal streaming recognizer is a prefix predictor, not a
transcriber.} Theorem~\ref{thm:maximal} says that at time $t$ the longest target
that is safe to commit and that the model has itself realized is $L_t$, the
longest earlier output of the model that is wholly a prefix of $\bar y$. The
objective is therefore $x_{\le t} \mapsto L_t$, a prediction of \emph{what the
final transcript will begin with}, not $x_{\le t} \mapsto y_t$, a transcription
of the clip. The two differ in both directions. $y_t$ normally carries a tail
beyond $L_t$, a completion guessed from a clip that ends mid-phrase, which later
audio is free to revise and which an append-only system must therefore not emit;
and when the current clip output is shorter than an earlier fully agreeing one,
$y_t$ is shorter than $L_t$ instead (Proposition~\ref{prop:incomparable}, second
witness). The relation between the two is a comparison, not a decomposition.
Training on $y_t$, or on the words of $\bar y$ whose audio has already ended
(typically longer than $L_t$ by the right-context lag, though not guaranteed to
be, since $L_t$ may contain a word whose audio has not yet ended), teaches the
model to emit content that the commit rule would make permanent, whatever later
audio shows.

%\bibliographystyle{plainnat}
%\bibliography{stable_prefix_intro_section}

\section{Training Configuration}
\label{app:configurations}

The tables below summarize the training corpus, augmentation,
SFT and GRPO settings.

\begin{table}[!ht]
\centering
\begin{tabularx}{\linewidth}{p{5.2cm}X}
\toprule
Setting & Value \\
\midrule
Base model & Qwen3-ASR-1.7B \\
Post-training data source & Proprietary; no public datasets used for SFT or GRPO \\
Training utterances & $600{,}000$: $300{,}000$ Chinese and $300{,}000$ English \\
Training audio & $1{,}500$ hours \\
Fbank feature hop & 10\,ms; 128 log-Mel coefficients \\
Encoder-frame spacing & $\delta=80\,\mathrm{ms}$ (12.5\,Hz), after $8\times$ CNN subsampling \\
Teacher annotation spacing & $\delta=80\,\mathrm{ms}$ \\
Annotation update & Forward LSP update with nondecreasing stored-prefix length \eqref{eq:lsp-target} \\
SFT audio attention & Dynamic windows from 1 to 8\,s \\
SFT example sampling & 10\% complete utterances; 90\% prefix cuts \\
SFT prefix-cut selection & Conditional on a prefix cut: 50\% within LSP agreement intervals, 50\% in their gaps; timestamps may fall between annotation endpoints \\
SFT additive-noise SNR & 10--35\,dB \\
SFT pure / mixed noise & 10\% pure-noise replacement; 30\% conditional mixed-noise probability \\
SFT reverberation & 40\%; 90\% real and 10\% simulated impulse responses \\
SFT language-tag flip & 10\% \\
\bottomrule
\end{tabularx}
\caption{Training corpus and data augmentation configuration.}
\label{tab:documented-configuration}
\end{table}

Table~\ref{tab:sft-run-settings} records the supervised
fine-tuning settings. The corpus, augmentation, and endpoint
sampling settings are listed in
Table~\ref{tab:documented-configuration}. GRPO is initialized from the SFT model.
Table~\ref{tab:grpo-run-settings} records the
optimization and rollout settings. The global batch and
policy minibatch count prompt groups; the GPU microbatch
counts individual trajectories.

\begin{table}[!ht]
\centering
\small
\begin{tabularx}{\linewidth}{p{4.5cm}X}
\toprule
Setting & SFT run \\
\midrule
Initialization & Qwen3-ASR-1.7B \\
Objective & Cross-entropy on LSP transcription and \waittoken{} targets \\
Optimizer & Fused AdamW \\
Learning rate & $10^{-5}$ \\
LR schedule & Linear decay after 2\% warmup \\
Epochs & 1 \\
Per-GPU batch / accumulation & 4 examples / 16 steps \\
Effective global batch & 64 examples \\
Audio encoder and projector & Frozen; rationale in Section~\ref{sec:training} \\
Trainable modules & Language-decoder parameters only \\
LoRA & Disabled \\
Dynamic audio attention & Enabled; 1--8\,s windows \\
\bottomrule
\end{tabularx}
\caption{SFT settings.}
\label{tab:sft-run-settings}
\end{table}

\begin{table}[!htbp]
\centering
\small
\begin{tabularx}{\linewidth}{p{5.0cm}X}
\toprule
Setting & GRPO run \\
\midrule
Backbone & Qwen3-ASR-1.7B \\
Initialization & Model produced by the SFT stage \\
Audio encoder and projector & Frozen \\
Trainable modules & Language-decoder parameters only \\
Policy optimization & GRPO; 1 optimization epoch per rollout batch \\
Optimizer & AdamW; weight decay 0.01 \\
Learning rate & $5\times10^{-6}$; constant, no warmup \\
Global batch & 64 prompt groups \\
Samples per prompt & 8; 512 trajectories per batch \\
Policy (PPO) minibatch & 64 prompt groups \\
Per-GPU microbatch & 1 trajectory \\
Policy clipping & 0.2 \\
KL regularization & Coefficient 0.1; \texttt{low\_var\_kl} estimator \\
Gradient clipping & Maximum norm 1.0 \\
Rollout sampling & Temperature 0.8; top-$p$ 0.95 \\
Prompt token limit & 4,096 \\
Generation limit & 32 tokens per turn \\
Cut-marker reward token & \waittoken{} \\
\bottomrule
\end{tabularx}
\caption{GRPO optimization and rollout settings. A trajectory may
span multiple streaming turns; the generation limit applies
to each turn.}
\label{tab:grpo-run-settings}
\end{table}

The GRPO reward combines the eight components in Table~\ref{tab:reward}. Details are as follows.

\begin{table}[!htbp]
\centering
\small
\begin{tabularx}{\linewidth}{@{}lrX@{}}
\toprule
Component & Weight & Role \\
\midrule
CER & 0.300 & Normalized character-level recognition accuracy \\
Precision & 0.150 & Edit-distance precision surrogate \\
Length & 0.150 & Excess output length \\
Repetition & 0.125 & Repeated runs and character patterns \\
Language Consistency & 0.150 & Chinese/Latin character composition \\
Tail & 0.050 & Excess trailing text inside LSP agreement intervals \\
Word Accuracy & 0.050 & Local recognition inside LSP agreement intervals \\
Cut Marker & 0.025 & Required \waittoken{} presence and placement \\
\bottomrule
\end{tabularx}
\caption{The eight reward components and their weights.
The weighted sum is used only after the special-case rules.}
\label{tab:reward}
\end{table}

At a sampled supervision time $t$, let $y_t$ be the generated decoder
response and $y_t^*$ its training target. 

\paragraph{Global Recognition Scores.}
Let $p(y_t,y_t^*)=\max(0,\max(|y_t|,|y_t^*|)-d(y_t,y_t^*))/|y_t|$ for $|y_t|>0$, where $|\cdot|$ counts characters and $d(\cdot,\cdot)$ is the character-level
Levenshtein distance.
The CER and edit-distance precision rewards are
\begin{equation*}
\begin{aligned}
r_{\mathrm{cer}}&=
\begin{cases}
  1,&y_t=y_t^*=\varepsilon,\\
  -1,&\text{exactly one of }y_t,y_t^*\text{ is empty},\\
  1-2\min\!\left(\dfrac{d(y_t,y_t^*)}{\max(|y_t^*|,1)},1\right),&\text{otherwise},
\end{cases}\\[4pt]
r_{\mathrm{prec}}&=
\begin{cases}
  1,&y_t=y_t^*=\varepsilon,\\
  0,&y_t=\varepsilon,\ y_t^*\neq\varepsilon,\\
  -1,&y_t\neq\varepsilon,\ y_t^*=\varepsilon,\\
  2p(y_t,y_t^*)-1,&\text{otherwise}
\end{cases},
\end{aligned}
\end{equation*}
where $\varepsilon$ is empty text.

\paragraph{Length and Repetition.}
For $y_t^*\neq\varepsilon$, let $\rho=|y_t|/|y_t^*|$. Then
\begin{equation*}
r_{\mathrm{len}}=
\begin{cases}
  -\min(|y_t|/10,1),&y_t^*=\varepsilon,\\
  0,&y_t^*\neq\varepsilon,\ \rho\leq1.5,\\
  -0.3(\rho-1.5)/1.5,&y_t^*\neq\varepsilon,\ 1.5<\rho\leq3,\\
  -0.5,&y_t^*\neq\varepsilon,\ \rho>3.
\end{cases}
\end{equation*}
Let $c$ be the length of the longest run of an identical character in $y_t$ (zero for
empty text), and let $P(y_t)$ indicate that some character $k$-gram,
$2\leq k\leq10$, occurs at least four times consecutively. Repetition is
\begin{equation*}
\begin{aligned}
 \phi(c)&=\begin{cases}
   -0.5,&c>10,\\
   -0.3,&5<c\leq10,\\
   0,&c\leq5,
 \end{cases}\\
 r_{\mathrm{rep}}&=\max\!\left(-1,\phi(c)-0.4\,\mathbf{1}[P(y_t)]\right).
\end{aligned}
\end{equation*}

\paragraph{Language Consistency.}
Let $\pi_{\mathrm{cjk}}(y_t)$ be the fraction of CJK characters among CJK
and Latin letters, set to zero if no such letters occur. With
$\eta=|\pi_{\mathrm{cjk}}(y_t)-\pi_{\mathrm{cjk}}(y_t^*)|$,
\begin{equation*}
r_{\mathrm{lang}}=
\begin{cases}
  0,&\eta=0\text{ or either text has no such letters},\\
  -\min(\eta/0.5,1),&\text{otherwise}.
\end{cases}
\end{equation*}

\paragraph{Tail and Local Word Accuracy.}
Writing $\Delta_{\mathrm{len}}=|y_t|-|y_t^*|$, the Tail reward is
\begin{equation*}
r_{\mathrm{tail}}=-0.3\,q(t)\,
 \mathbf{1}[\Delta_{\mathrm{len}}>3]\,
 \min\!\left(\frac{\Delta_{\mathrm{len}}}{10},1\right),
\end{equation*}
where $q(t) = \mathbf{1}\!\left[\exists i:\ t \in I_i\right]$.
For Word Accuracy, tokenize $y_t,y_t^*$ into $W,W^{\ast}$, grouping consecutive
ASCII alphanumeric characters and treating Chinese characters separately.
Let $\kappa=|\operatorname{lcp}(W,W^{\ast})|$. If both sequences continue
past $\kappa$, let $w,w^{\ast}$ be their first divergent tokens. Word Accuracy
scores this local agreement rather than the accuracy of the entire
transcript:

\begin{equation*}
r_{\mathrm{wacc}}=q(t)
\begin{cases}
  0,&\kappa=0,\\
  1,&\kappa>0,\ \text{both sequences end at }\kappa,\\
  -1,&\kappa>0,\ W^{\ast}\text{ ends but }W\text{ continues},\\
  -0.5,&\kappa>0,\ W\text{ ends but }W^{\ast}\text{ continues},\\
  1-2\min\!\left(\dfrac{d(w,w^{\ast})}{\max(|w|,|w^{\ast}|)},1\right),
    &\text{otherwise}.
\end{cases}
\end{equation*}

\paragraph{Cut Marker.}
Let $b_{\mathrm{end}}(y_t)=\mathbf{1}[y_t\text{ ends with }\waittoken{}]$ and
$b_{\mathrm{mid}}(y_t)=\mathbf{1}[\waittoken{}\text{ occurs in a nonterminal position of }y_t]$.
The marker score and penalty are
\begin{equation*}
\begin{aligned}
s_{\mathrm{cut}}&=
\begin{cases}
  1,&b_{\mathrm{end}}(y_t)=b_{\mathrm{end}}(y_t^{\ast})=1,\\
  0,&b_{\mathrm{end}}(y_t^{\ast})=0\text{ and }\waittoken{}\notin y_t,\\
  -0.5,&\text{otherwise},
\end{cases}\\[3pt]
r_{\mathrm{cut}}&=\max(-1,s_{\mathrm{cut}}-0.3\,b_{\mathrm{mid}}(y_t)).
\end{aligned}
\end{equation*}

\paragraph{Combination and Ordered Overrides.}
With weights $w_k$ from Table~\ref{tab:reward},
\begin{equation*}
 R_{\mathrm{weighted}}=\sum_{k=1}^{8}w_k r_k.
\end{equation*}
The raw reward uses the following branches in the order shown:
\begin{equation*}
R_{\mathrm{raw}}=
\begin{cases}
  1,&y_t=y_t^{\ast}=\varepsilon,\\
  -0.5-0.5\min(|y_t|/50,1),&y_t^{\ast}=\varepsilon,\ y_t\neq\varepsilon,\\
  1,&y_t=y_t^*,\ b_{\mathrm{end}}(y_t)=b_{\mathrm{end}}(y_t^{\ast}),\\
  -0.7,&y_t^{\ast}\neq\varepsilon,\ y_t=\varepsilon,\\
  R_{\mathrm{weighted}},&\text{otherwise}.
\end{cases}
\end{equation*}
\section{Evaluation Details}
\subsection{Datasets and Baselines}
\label{app:evaluation-data}

The main evaluation comprises eight English test sets: AMI
\citep{carletta2006ami}, Giga-clean \citep{chen2021gigaspeech},
LibriSpeech [clean, other] \citep{panayotov2015librispeech},
VoxPopuli \citep{wang2021voxpopuli}, Earnings22
\citep{delrio2022earnings22}, TED-LIUM
\citep{hernandez2018tedlium3}, and RealSI. The five Chinese
test sets are WenetSpeech [net, meeting]
\citep{zhang2022wenetspeech}, SPEECHIO [06, 07]
\citep{speechio_asr_benchmark}, and RealSI. RealSI contains
long-form audio from real-world simultaneous interpretation across
ten topics
\citep{cheng2024clasi}. We transcribe its English and Chinese source
recordings. These cohorts are used throughout the main tables and
figures. Transfer is evaluated separately on 18 test sets in 10
additional languages from FLEURS \citep{conneau2023fleurs},
Multilingual LibriSpeech (MLS) \citep{pratap2020mls}, Common Voice
\citep{ardila2020commonvoice}, and GigaSpeech2
\citep{yang2025gigaspeech2}, and these scores do not enter the
English or Chinese macro averages.
R2T2 uses only proprietary data for SFT and GRPO, with none of the
evaluation datasets included. However, the base model may have encountered
some of these datasets during pre-training.
We compare R2T2 with Commercial A, Commercial B, and AssemblyAI,
as well as the open-source models Qwen3-ASR-1.7B, X-ASR, Nemotron 3.5,
Voxtral Realtime, and WhisperRT
\citep{shi2026qwen3asr,xasr_modelcard,nvidia2026nemotron35asr,mistral2026voxtralmodelcard,krichli2026whisperrt}.
We distinguish incremental audio processing from the policy for
revising or committing output.

\subsection{Cross-Language Recognition Results}
\label{app:language-transfer}

Table~\ref{tab:language-transfer} reports recognition error for
the 18 test sets in 10 languages absent from R2T2 post-training.
Japanese, Korean, and Thai use CER, and the other languages use
WER. All R2T2 settings use the same checkpoint. Full context
denotes decoding each complete utterance once. Streaming columns
specify the chunk duration and provisional-token budget $U$.
Scores are reported for each test set without hallucination filtering.
The Average row gives equal weight to each of the 18 test sets,
combining their WER or CER percentages rather than pooling edit counts.

\begin{table}[!htbp]
\centering
\small
\setlength{\tabcolsep}{5pt}
\renewcommand{\arraystretch}{1.10}
\begin{tabular}{@{}l*{5}{r}@{}}
\toprule
Test set & \multicolumn{3}{c}{Qwen3-ASR-1.7B} & \multicolumn{2}{c}{R2T2} \\
\cmidrule(lr){2-4}\cmidrule(lr){5-6}
& Full context & 2\,s & 160\,ms & 160\,ms & 160\,ms \\
& & $U=5$ & $U=1$ & $U=0$ & $U=1$ \\
\midrule
Arabic: FLEURS & 9.53 & 11.32 & 56.63 & 14.86 & 13.52 \\
\addlinespace[1pt]
German: FLEURS & 3.95 & 10.77 & 36.49 & 12.35 & 8.05 \\
German: MLS & 6.99 & 7.55 & 26.83 & 12.09 & 9.52 \\
\addlinespace[1pt]
Spanish: FLEURS & 2.63 & 3.55 & 31.00 & 5.22 & 3.97 \\
Spanish: MLS & 3.81 & 3.80 & 18.36 & 5.45 & 4.47 \\
\addlinespace[1pt]
French: FLEURS & 4.30 & 5.67 & 39.61 & 7.65 & 6.35 \\
French: MLS & 5.09 & 5.43 & 22.12 & 6.90 & 6.14 \\
\addlinespace[1pt]
Japanese: FLEURS & 5.42 & 7.23 & 36.07 & 10.15 & 8.66 \\
\addlinespace[1pt]
Korean: FLEURS & 4.22 & 6.20 & 40.66 & 6.77 & 7.52 \\
\addlinespace[1pt]
Portuguese: FLEURS & 4.47 & 5.02 & 23.94 & 8.02 & 6.76 \\
Portuguese: MLS & 6.43 & 6.65 & 20.94 & 8.64 & 7.41 \\
\addlinespace[1pt]
Russian: FLEURS & 4.94 & 6.95 & 53.92 & 11.26 & 8.26 \\
\addlinespace[1pt]
Thai: Common Voice & 3.50 & 3.62 & 24.12 & 5.58 & 5.18 \\
Thai: FLEURS & 6.81 & 8.40 & 21.25 & 10.14 & 9.83 \\
Thai: GigaSpeech2 & 12.39 & 12.97 & 18.43 & 15.39 & 14.93 \\
\addlinespace[1pt]
Vietnamese: Common Voice & 8.48 & 8.51 & 38.92 & 11.87 & 11.25 \\
Vietnamese: FLEURS & 6.99 & 8.66 & 28.11 & 11.88 & 10.89 \\
Vietnamese: GigaSpeech2 & 7.85 & 7.83 & 16.04 & 8.65 & 8.42 \\
\midrule
Average & 5.99 & 7.23 & 30.75 & 9.60 & 8.40 \\
\bottomrule
\end{tabular}
\caption{Cross-language recognition error (\%) for Qwen3-ASR-1.7B and R2T2.
Japanese, Korean, and Thai use CER; all other languages use WER.}
\label{tab:language-transfer}
\end{table}

\end{document}